\pdfoutput=1
\documentclass[11pt]{article}

\usepackage{acl}

\usepackage{times}
\usepackage{latexsym}

\usepackage[T1]{fontenc}
\usepackage[utf8]{inputenc}

\usepackage{microtype}

\usepackage{inconsolata}

\usepackage{algorithm}
\usepackage{algorithmic}

\usepackage{amsmath}
\usepackage{todonotes}
\usepackage{booktabs}
\usepackage{subcaption}
\usepackage{graphicx}

\usepackage{newfloat}
\usepackage{listings}
\usepackage{colortbl}
\usepackage{multirow}

\definecolor{'deep1'}{HTML}{C5E6F8} 
\definecolor{'shallow1'}{HTML}{E4F3FC} 
\definecolor{'deep2'}{HTML}{E5F5B7} 
\definecolor{'shallow2'}{HTML}{F3FADF}

\definecolor{'wit'}{HTML}{FBFBFB}
\definecolor{'gry'}{HTML}{EEEEEE}

\title{How Proficient Are Large Language Models in Formal Languages? \\ An In-Depth Insight for Knowledge Base Question Answering}

\author{
\textbf{Jinxin Liu}$^1$, \textbf{Shulin Cao}$^1$, \textbf{Jiaxin Shi}$^3$, \textbf{Tingjian Zhang}$^1$, \\
\textbf{Lunyiu Nie}$^1$, \textbf{Linmei Hu}$^4$, \textbf{Lei Hou}$^1$\thanks{\quad Corresponding author: houlei@tsinghua.edu.cn}, \textbf{Juanzi Li}$^{1,2}$\\
  $^1$ Department of Computer Science and Technology, BNRist; \\
  $^2$ THU - Siemens Ltd., China Joint Research Center for Industrial Intelligence and IoT;\\ Tsinghua University, Beijing, China \\ 
  $^3$ Huawei Cloud Computing Technologies Co., Ltd. \\
  $^4$ School of Computer Science and Technology, Beijing Institute of Technology \\
  \texttt{liujinxi20@mails.tsinghua.edu.cn} \\}

\begin{document}
\maketitle
\begin{abstract}
%Recent advancement in the capabilities of large language models (LLMs) has triggered a new surge in LLMs' evaluation. Most recent evaluation works tends to evaluate the comprehensive ability of LLMs over series of tasks. However, the deep structure understanding of natural language is rarely explored. In this work, we examine the ability of LLMs to deal with structured semantics on the tasks of question answering with the help of the human-constructed formal language. Specifically, we implement the inter-conversion of natural and formal language through in-context learning of LLMs to verify their ability to understand and generate the structured logical forms. Extensive experiments with models of different sizes and in different formal languages show that today's state-of-the-art LLMs' understanding of the logical forms can approach human level overall, but there still are plenty of room in generating correct logical forms, which suggest that it is more effective to use LLMs to generate more natural language training data to reinforce a small model than directly answering questions with LLMs. Moreover, our results also indicate that models exhibit considerable sensitivity to different formal languages. In general, the formal language with the lower the formalization level, i.e. the more similar it is to natural language, is more LLMs-friendly.

%Recent advancement in the capabilities of large language models (LLMs) has triggered a re-structuring of traditional natural language tasks, including 
Knowledge Base Question Answering (KBQA) aims to answer natural language questions based on facts in knowledge bases.
A typical approach to KBQA is semantic parsing, which translates a question into an executable logical form in a formal language.
Recent works leverage the capabilities of large language models (LLMs) for logical form generation to improve performance. 
However, although it is validated that LLMs are capable of solving some KBQA problems, there has been little discussion on the differences in LLMs' proficiency in formal languages used in semantic parsing.
%The \textcolor{red}{general approach} to KBQA in recent \textcolor{red}{years} is semantic parsing, which translate questions into executable logical forms. 
%Due to the rapidly risen capability of large language models (LLMs), most recent works have begun to leverage LLMs and shown promising results. However, few has discussed \textcolor{red}{the influence of different formal language on the performance }and how well LLMs master them. 
In this work, we propose to evaluate the understanding and generation ability of LLMs to deal with differently structured logical forms by examining the inter-conversion of natural and formal language through in-context learning of LLMs. Extensive experiments with models of different sizes show that state-of-the-art LLMs can understand formal languages as well as humans, but generating correct logical forms given a few examples remains a challenge.
%\textcolor{red}{today's state-of-the-art LLMs' understanding of logical forms approach to human, but generation with only a few example still is a challenge.}
Most importantly, our results also indicate that LLMs exhibit considerable sensitivity. In general, the formal language with a lower formalization level, i.e., the more similar it is to natural language, is more friendly to LLMs. Code and data can be found at \url{https://github.com/Matthewlliu/structure_probe}.

\end{abstract}

\section{Introduction}
\label{sec:intro}

%The recent advancement of large language models (LLMs), showcasing their remarkable language understanding and generation capabilities~\cite{DBLP:journals/corr/abs-2303-12712}, has garnered significant attention from researchers in artificial intelligence. It has also triggered a surge in LLMs' evaluation endeavors aimed at exploring the boundaries of LLMs' capabilities.
%There have been numerous works to probe large language models with varying sizes such as BERT~\cite{DBLP:conf/naacl/DevlinCLT19} and GPT~\cite{radford2018improving}. 
%Previous works that probe large language models with varying sizes such as BERT~\cite{DBLP:conf/naacl/DevlinCLT19} and GPT~\cite{radford2018improving} mainly focus on probing models' linguistic knowledge~\cite{DBLP:conf/naacl/HewittM19, DBLP:conf/blackboxnlp/ClarkKLM19, DBLP:conf/naacl/Liu0BPS19, DBLP:conf/acl/WuCKL20} and world knowledge~\cite{DBLP:conf/emnlp/PetroniRRLBWM19, DBLP:conf/acl/DaiDHSCW22, DBLP:journals/tacl/JiangXAN20, DBLP:conf/naacl/ZhongFC21}. Most recent works tend to evaluate models' comprehensive capacities over series of tasks~\cite{DBLP:journals/corr/abs-2302-04023, DBLP:journals/corr/abs-2306-09296} such as reasoning and interactivity.

Knowledge Base Question Answering (KBQA) is a challenging natural language processing (NLP) task to answer natural language questions based on fact triples stored in the knowledge base (KB), such as Wikidata~\cite{DBLP:journals/cacm/VrandecicK14} and Freebase~\cite{DBLP:conf/sigmod/BollackerEPST08}.
%Since the emergence of language models like BERT~\cite{DBLP:conf/naacl/DevlinCLT19} and GPT~\cite{radford2018improving}, single-relation factoid questions have begun to be trivial~\cite{DBLP:conf/emnlp/PetrochukZ18}, and researchers turn to solve more complex questions that requires compositional reasoning skills, such as multi-hop inference and quantitative comparison~\cite{DBLP:conf/naacl/TalmorB18}.
In recent years, a typical paradigm of KBQA methods is semantic parsing~\cite{DBLP:conf/emnlp/BerantCFL13, DBLP:conf/acl/CaoZLLY19, DBLP:conf/acl/YeYHZX22, DBLP:journals/corr/abs-2210-12925}, where natural language questions (NLQs) are translated into their corresponding structured logical forms (LFs), such as KoPL~\cite{DBLP:conf/acl/CaoSPNX0LHZ22}, SPARQL~\cite{DBLP:conf/semweb/PerezAG06} or Lambda DCS~\cite{DBLP:journals/corr/Liang13}. The logical forms are capable of expressing multiple reasoning operations such as multi-hop inference and quantitative comparison, and can be executed on KBs to get accurate answers.

The recent advancements of large language models (LLMs)~\cite{DBLP:journals/corr/abs-2303-08774} have led to significant attention on utilizing LLMs for KBQA. 
Previous works have validated the ability of LLMs to memorize, understand and apply knowledge for reasoning~\cite{DBLP:journals/corr/abs-2306-09296}, and serve as agents to solve KBQA problem~\cite{DBLP:journals/corr/abs-2308-03688}. 
Additionally, others introduced many techniques to improve the performance, such as in-context learning~\cite{DBLP:journals/corr/abs-2305-01750}, chain-of-thought~\cite{DBLP:conf/emnlp/LiangWZWQL23}, and instruction-tuning~\cite{DBLP:journals/corr/abs-2310-08975}.

%Previous works such as~\cite{DBLP:journals/corr/abs-2305-01750} leveraged the in-context learning ability of LLMs to do parsing by providing a few NLQ-LF pairs. Others~\cite{DBLP:journals/corr/abs-2310-08975} proposed combining LLMs' instruction fine-tuned generation with retrieval of entities and relations.

However, there has been little discussion on the differences in the proficiency of LLMs in different formal languages that used as paring targets in semantic parsing.
It is proved that LLMs can do well on programming language such as Python~\cite{DBLP:conf/icml/GaoMZ00YCN23} with sufficient data in the pretraining, but it remains intriguing that how well do LLMs master other formal languages without extra data and further fine-tuning. 
By examining the proficiency of LLMs in formal languages, we can gain a better understanding of LLMs' upper limits.
%when training them for semantic parsing or other purpose. 
It also would be advantageous to choose appropriate models and formal languages for specific scenarios if different LLMs have varying levels of proficiency in different formal languages.
In this paper, we propose to evaluate the inherent understanding and generation ability of the formal language in the original LLMs without additional fine-tuning. 
We define two evaluation tasks based on sub-tasks of KBQA: 1) \textbf{Formal Language Understanding}, which aims to translate a LF into its corresponding NLQ. The translation process can be considered as the model interpreting the provided LFs in natural language, demonstrating LLMs' understanding ability of formal language; 
2) \textbf{Formal Language Generation}, which aims to correctly convert a NLQ into its corresponding LF, requiring the model to not only understand but also generate LFs, demonstrating its capability in generation. 

% Naturally, this ability is evaluated by examining the twofold performance on both understanding and generation of LLMs.

% Today, different formal languages, which are usually defined by a group of identifiers with straightforward grammar and mathematical attributes, are vastly used for various tasks, e.g., Lambda Calculus~\cite{artzi2013semantic}, Lambda DCS~\cite{liang2013lambda}, etc.
% To design the probing framework, we 
With respect to the formal languages for evaluation, according to the varying levels of formalization (may be broadly understood as the dissimilarity to natural language, i.e. the higher the level of formalization, the less similar it is to natural language) and different logical structures (e.g. tree, graph or chain), we choose Lambda DCS~\cite{DBLP:journals/corr/Liang13}, SPARQL~\cite{DBLP:conf/semweb/PerezAG06}, and KoPL~\cite{DBLP:conf/acl/CaoSPNX0LHZ22} as representative formal languages, which are commonly used for knowledge based question answering research~\cite{DBLP:conf/emnlp/NieCSST0LZ22, DBLP:conf/acl/YeYHZX22, DBLP:conf/emnlp/ShinLTCRPPKED21}. 

For the generation methods, to reflect the inherent proficiency of LLMs, 
%we employ methods not requiring a large amount of additional data.Specifically, 
we combine the in-context learning ability~\cite{DBLP:conf/nips/BrownMRSKDNSSAA20} of LLMs and chain-of-thought generation~\cite{DBLP:conf/nips/Wei0SBIXCLZ22} for both evaluation tasks, where the desired outputs are generated conditioned on the input along with a few demonstration pairs of NLQs and LFs carefully selected from a seed dataset. For demonstration selection, to ensure that the logical structure of the examples should be as similar as possible to the target, we carefully design a greedy search algorithm based on the minimum edit distance to solve a maximum coverage problem. 

For the quality evaluation of the generated NLQ, to avoid the inaccuracy of automatic metrics and the labor-intensive human evaluation, we propose a contrastive evaluation approach. This involves separately training a semantic parser using LLM-generated data and comparing them with parser trained using manually labeled data. By comparing the performance of the parsers, we can assess the differences in quality between the data generated by the LLMs and the manually labeled data.
%design multiple search strategies for different formal languages and different tasks, 

% In summary, this article presents a comprehensive exploration of LLMs' performance in two sub-tasks, corresponding to their capabilities in understanding and generating structured semantics. Starting from different models, knowledge bases, and formal languages, we propose a baseline method based on retrieval-based in-context learning. 

Our findings indicate that 
%there is still a gap between LLMs and human in terms of structured \textcolor{red}{semantics} understanding.
%\textcolor{red}{Coinciding with our intuition, 
%the generation capability of LLMs for structured semantics is much weaker than their understanding ability. 
LLMs have approached the human annotators in the task of formal language understanding that generate natural language questions from logical forms. However, conversely, challenges still exist in the task of formal language generating if only a few examples are given.
Importantly, we observe that models exhibit the sensitivity to different logical forms. Overall, the lower the level of formalization (the similar it is to natural language), the easier it is for models to understand and generate. 
%These findings suggest the feasibility of employing LLMs combining with KBs to tackle the complex reasoning that currently still \textcolor{red}{pose} a challenge~\cite{DBLP:journals/corr/abs-2302-04023}.
In conclusion, this study examines the proficiency of LLMs in formal language understanding and generation, and helps to provide valuable insights for LLMs-based reasoning approaches.

\section{Related Work}
\noindent \textbf{Knowledge Base Question Answering.} Typical methods for solving KBQA problems can be broadly divided into two categories. One category is the retrieval-based method. These methods usually directly output the answer by retrieving triples and subgraphs that related to the question from KB or embedded memory~\cite{DBLP:conf/emnlp/SunBC19, DBLP:conf/emnlp/ShiC0LZ21, DBLP:conf/acl/ZhangZY000C22,DBLP:conf/naacl/OguzCKPOSGMY22, DBLP:conf/cikm/DongLWZXX23}. Another is the semantic-parsing-based method, which translates questions into logical forms executable against KBs. The logical forms are usually generated by step-by-step graph searching and generation~\cite{DBLP:conf/www/GuKVSLY021,DBLP:conf/iclr/JiangZ0W23,DBLP:conf/emnlp/JiangZDYZW23,DBLP:conf/acl/Gu0023} or by sequence-to-sequence model that trained with parallel data~\cite{DBLP:conf/acl/YeYHZX22, DBLP:conf/acl/CaoS0LY0L0X22, DBLP:journals/dint/DuLWCY22, DBLP:journals/corr/abs-2210-12925, DBLP:conf/iclr/YuZNZL0HWWX23, DBLP:journals/corr/abs-2310-08975}.

Since the logical form can facilitate communication between the model and the KB, the latter category usually out-performance the former and also enjoy a better interpretability. Therefore, our work sets out to explore the role of different formal languages in KBQA in the era of LLMs.

\noindent \textbf{Knowledge Base Question Generation.} Early KBQG works mainly utilize templates to generate questions. These approaches~\cite{DBLP:conf/emnlp/BerantCFL13, DBLP:conf/naacl/TalmorB18, DBLP:conf/ictir/SeylerYB17} usually leverage search engines or manual design to obtain the templates. However, template-based methods are restricted by heavy manual labor. With the advancement of deep neural networks, neural-based methods~\cite{DBLP:conf/naacl/ElSaharGL18, DBLP:conf/eacl/KhapraRJR17} are proposed to generate questions automatically by the sequence-to-sequence model. Since the semantic information in the structure of the entity graph is considered useful for KBQG, graph neural models~\cite{DBLP:conf/semweb/KumarHRQGL19, DBLP:journals/corr/abs-2004-06015, DBLP:conf/emnlp/Guo0WZ0022} are applied to generate multi-hop questions based on subgraphs. Most recently, \citeauthor{DBLP:conf/emnlp/LiangWZWQL23} utilize LLMs with its few-shot learning ability and chain-of-thought reasoning, proposing to generate question directly from its formal language. In the formal language understanding task of our work, we are inspired by the idea of \citeauthor{DBLP:conf/emnlp/LiangWZWQL23} and customize our own chain-of-thought methods.

\noindent \textbf{Evaluation of LLMs.} From the advent of pretrained language models (PLMs) such as BERT~\cite{DBLP:conf/naacl/DevlinCLT19} and GPT~\cite{radford2018improving} to the emergence of increasingly larger and powerful LLMs~\cite{DBLP:conf/nips/BrownMRSKDNSSAA20, DBLP:journals/jmlr/ChowdheryNDBMRBCSGSSTMRBTSPRDHPBAI23,DBLP:journals/corr/abs-2211-05100,DBLP:conf/iclr/ZengLDWL0YXZXTM23, DBLP:journals/corr/abs-2307-09288} in recent years, language models have changed the paradigms of many traditional task. At the same time, the evaluations of language models are also ongoing.

Early works explored PLMs' capability boundaries including linguistics knowledge~\cite{DBLP:conf/naacl/HewittM19, DBLP:conf/blackboxnlp/ClarkKLM19, DBLP:conf/naacl/Liu0BPS19} as well as world knowledge like entities~\cite{DBLP:journals/corr/abs-2003-05473}, relations~\cite{DBLP:conf/emnlp/PetroniRRLBWM19, DBLP:journals/tacl/JiangXAN20, DBLP:conf/naacl/ZhongFC21}, and concepts~\cite{DBLP:conf/emnlp/PengWHJ0L0022, DBLP:conf/iclr/DalviKAD0S22}. Recent works have included comprehensive tasks and datasets to create new benchmarks for LLMs~\cite{DBLP:journals/corr/abs-2302-04023, DBLP:journals/corr/abs-2206-04615, DBLP:journals/corr/abs-2306-09296, DBLP:journals/corr/abs-2308-03688}. Our work are inspired by above studies, extending the evaluation of LLMs from the natural language domain to various formal languages.

\begin{figure*}[h]
    \centering
        \includegraphics[width=0.8\textwidth]{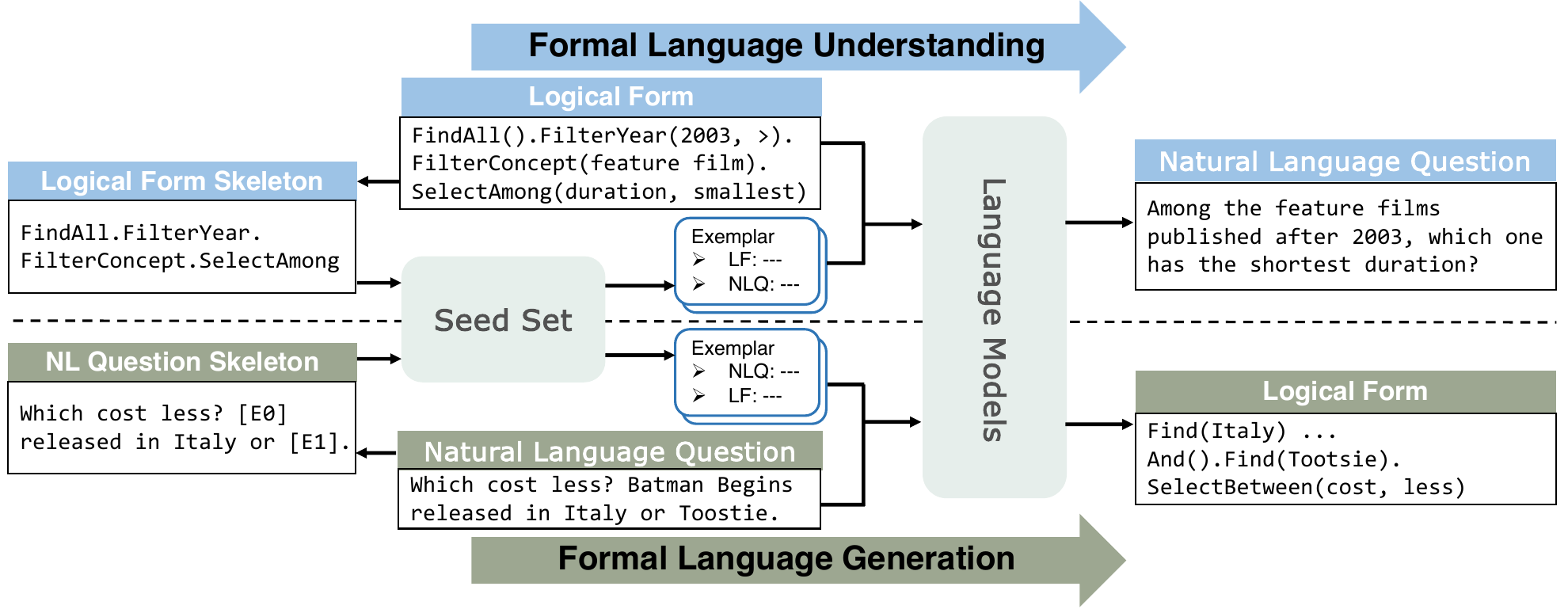}
        \caption{ A simple illustration for the probing task of both formal language understanding and generation.}
     \label{fig:intro}
\end{figure*}

\section{Evaluation Task Definition}
\label{sec: pef_eva}
As shown in Figure~\ref{fig:intro}, 
We define two probing tasks, namely the \textbf{formal language understanding} and \textbf{formal language generation}. In this section, we introduce the formalized definitions of these two tasks and how to assess LLMs' performance.
%the probing process for both tasks is literally the same. The basic technique is in-context learning, where we prompt the model by providing several example input-output pairs along with the target input, and content generated by the model is the output.

\subsection{Formal Language Understanding} The goal of the task is for a LLM $M$ to translate a LF input to its corresponding NLQ. Formally, we have a target set $\mathcal{T} = \{l*\}$ of LFs, and a seed set $\mathcal{S} = \{ (l, q) \}$ of LF - NLQ pairs. To assemble the demonstration, for every $l$ in $\mathcal{T}$ we need to retrieve $k$ pairs of LFs and NLQs $ (l_1, q_1), \cdots, (l_k, q_k) $ from $\mathcal{S}$. Conditioned on the examples and $l$, the model translates it into a NLQ.

%Evaluating the quality of the generated $q$, considering the sheer size of the dataset, is difficult to check every $q$'s correctness by human. 
For the evaluation of the quality of the generated NLQs, the commonly used automatic metrics to compare text similarity like BLEU~\cite{DBLP:conf/acl/PapineniRWZ02} and BERT-Score~\cite{DBLP:conf/iclr/ZhangKWWA20} are not reliable enough. 
%In this case, we choose to evaluate the generation quality of a model \textit{M} indirectly. We evaluate the quality of the model-generated data 
Instead, we evaluate the generation quality of a model \textit{M} indirectly
by comparing the performance of the parser trained on the model-generated data and the parser trained on the manually-labeled data. Formally, given the training set $\{(q*,l*)\}$, where the $l*$ is the LF and $q*$ is the corresponding human-labeled NLQ, we train a baseline semantic parser $P_{human}$. Then we take $\{l*\}$ as the target set $\mathcal{T}$, using $M$ to generate a same-size pseudo training set $\{(q_M, l*)\}$, which is used to train another parser $P_M$. In this case, the generation quality of $M$ is measured by $P_M$'s performance $Accuracy_{P_M}$ to $Accuracy_{P_{human}}$ of $P_{human}$. Higher score means better quality of the model-generated questions, indicating closer understanding ability of $M$ is to human. 

\subsection{Formal Language Generation} 
The goal is for a LLM $M$ to directly translate a NLQ back to its correct LF. Similarly, we have a target set of $\mathcal{T} = \{q*\}$ of NLQs, and a seed set $\mathcal{S} = \{(l, q) \}$ of LF - NLQ pairs. For every $q$ in $\mathcal{T}$, we retrieve $k$ pairs of NLQs and LFs $(q_1, l_1), \cdots, (q_k, l_k) $ from $\mathcal{S}$ to assemble the final prompt. The model is supposed to generate the correct LF $l$ conditioned on the examples and $q$.

 The evaluation of the generated $l$ is relatively easier. To evaluate whether the generated LF are correct and semantically equivalent to the input $q$, we can use the either the exact match score with the golden logical forms, or the accuracy of the answer by putting the logical forms into an executor.

\section{Formal Language and Datasets}
In this section, we will introduce the details of the formal languages and datasets tested in this work.
%of the probing \textcolor{red}{framework, which contains two parts: (1) the formal languages and their corresponding datasets, (2) the probing task design with the formal descriptions, and the evaluation of semantic correctness of the generated natural language or formal language text.}
%\subsection{Formal Language}

\begin{figure}[t]
    \centering
    \includegraphics[width=.48\textwidth]{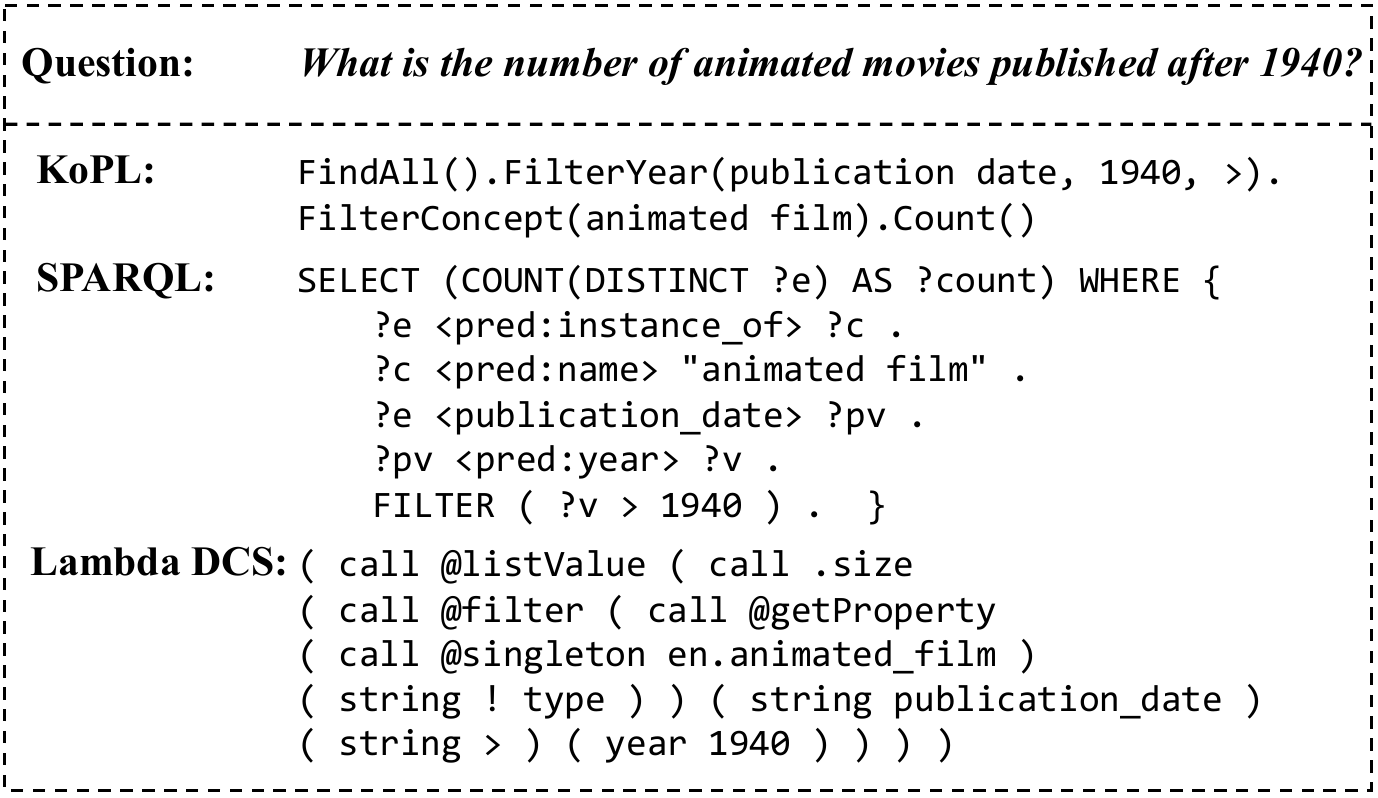}
    \caption{\label{fig:intro_b} An example of a natural language question and its corresponding logical forms in KoPL, SPARQL, and Lambda DCS.}
\end{figure}

As mentioned in Section~\ref{sec:intro}, we choose three representative formal languages according to the varying levels of formalization and different logical structure, and they are Lambda DCS, SPARQL, and KoPL. Some examples are shown in Figure~\ref{fig:intro_b}.

\noindent \textbf{Lambda DCS} is a tree-structured programming language developed from Lambda calculus, similar to church and s-expression. Lambda DCS removes the explicit variables in Lambda calculus, making it similar to dependency-based compositional semantics~\cite{DBLP:journals/corr/Liang13}. For this language, we use Overnight dataset~\cite{DBLP:conf/acl/WangBL15}, which contains over 13,000 data examples in eight domains extracted from Freebase. 
%Each data entry contains a logical form and a canonical utterance generated by domain-general grammar, a paraphrased natural language question and its answer.
We follow the standard split used in \citeauthor{DBLP:conf/acl/WangBL15}.

\noindent \textbf{SPARQL} is a popular query language and it provides a standardized way for users to search and retrieve information stored in RDF databases and other Linked Open Data\footnote{\url{https://www.w3.org/TR/sparql11-query/}}. 
%In SPARQL query, it describes the relations between entities mainly using triples which follow the ``subject-predicate-object'' pattern. Thus, it fundamentally describes a query statement in the form of a graph structure. 
The SPARQL describes the relations between entities using triples in the form of a graph structure.
For this language, we use the GrailQA dataset~\cite{DBLP:conf/www/GuKVSLY021}, which is constructed based on Freebase and comprises a total of over 50,000 data entries along with their entity linking results.
%The author also provided the entity linking results for the questions in the dataset, which also come in handy in our method. 
We also followed the standard split used by the author~\cite{DBLP:conf/www/GuKVSLY021}.

\noindent \textbf{KoPL}~\cite{DBLP:conf/acl/CaoSPNX0LHZ22} is a programming language
%that is both compositional and interpretable, designed to explicitly represent the reasoning process required to solve complex question~\cite{DBLP:conf/acl/CaoSPNX0LHZ22}. In this language, a program is 
constructed using symbolic functions, which define the fundamental and atomic operations performed on knowledge bases. These functions are combined according to the ``chain-of-thought'' of the reasoning process, forming a chain structure program. For this language, we use the KQA Pro dataset~\cite{DBLP:conf/acl/CaoSPNX0LHZ22}, which is based on Wikidata and comprises of over 100,000 data entries. Each data entry includes a NLQ along with its corresponding KoPL and SPARQL query. We followed the standard split described in \citeauthor{DBLP:conf/acl/CaoSPNX0LHZ22}.

Basic features of these formal languages can be concluded that (1) KoPL and Lambda DCS can both potentially better reflect the ``chain-of-thought'' reasoning process than SPARQL, and (2) KoPL is more well-modularized and uses more human-readable identifiers and function input, making it closer to the distribution of natural language.

\section{Implementation}
As mentioned above, we mainly leverage the in-context learning (ICL) ability of LLMs to generate the output for the probing task. The demonstration selection is considered as the most critical part of this method.
In this work, we adopt the principle to search most similar examples to the target $l$, and decently order the examples by the similarity~\cite{liu2021makes} in the prompt.
%so that the more similar example is more close to the input.
%Usually the demonstration examples are selected by the principle that the examples should be similar to the final target $l$, and there are a number of simple commonly used criteria for determining the similarity~\cite{liu2021makes, DBLP:conf/acl/SorensenRRSRDKF22}. Another important aspect of demonstration learning is the order of the selected examples in the prompt~\cite{DBLP:conf/acl/LuBM0S22}. In this work, we adopt the simple solution of \citeauthor{liu2021makes}, which decently places the examples by the similarity to the input so that the more similar example is more close to the input.

\subsection{Formal Language Understanding}
%Formal Language Understanding is to probe the ability for LLMs to correctly translate a logical form representation into a natural language question, which is defined as the measure of LLMs' ability to understand structured semantics. 
In this task, the input of LLMs is the LF $l*$, so we search for examples $(l, q)$ from $\mathcal{S}$ where all $l$s are most similar to $l*$.

We consider that the retrieved examples should (1) have the most similar \textbf{logical structure} to the structure of the target logical form $l*$ and (2) share as many same \textbf{relations} as possible with $l*$.

\subsubsection{Structure-Preserving Principle}
In order to find the most structure-similar examples from $\mathcal{S}$, we first transform the original logical form $l*$ into a simple rooted tree-like structure $s*$ called skeleton, where $s*\leftarrow f(l*)$, f being the extraction function. Specifically, KoPL program is already a tree of functions, therefore the skeleton of KoPL is the tree formed by removing the functions' inputs. The Lambda DCS program is similar to KoPL, since it can be treated as a bracket tree. The SPARQL program is more complicated, since it depicts a graph by some triples. In this case, we use the corresponding S-expression program instead, which is also bracket tree. Afterwards, we group the examples in $\mathcal{S}$ using the skeleton of logical form as the key.

Then we find the most similar structure naturally by computing the tree edit distance (TED) between $s*$ and skeleton keys of $\mathcal{S}$. However, considering the overhead of the minimum TED algorithm, we serialized the tree structure and apply the simple minimum edit distance (ED) in practice. In general, these two algorithms can produce every different results. But due to the grammar restriction of program, the candidates at small distances computed by TED are almost the same to those of ED. For example, in KoPL there are some common fixed patterns like \textit{Find()} $\rightarrow$ \textit{Relate()} $\rightarrow$ \textit{Filter()}. 

\subsubsection{Content-Preserving Principle}
\label{sec:prin2}
The meaning of content here is two-fold. First off, there should be no symbols of $l*$ unseen in the demonstration examples.
%we do not want the target logical form $l$ to contain any unseen symbols of the formal language, 
Taking KoPL as example again, it means the function names need to be covered by demonstration examples as many as possible. This is a max cover problem and we perform a k-step greedy search based on the previous ranking result by edit distance. Specifically, providing there are $m$ skeletons $S = \{s_1, \cdots, s_m\}$ that are closest to the skeleton $s*$ of $l*$ at a distance of $d_0$, we select a $s_{t_i}$ at each time step $i$, so that,
\begin{equation}
\begin{aligned}
    s_{t_i} &= \arg\min\limits_{s\in S_i }{|s*|_i - | s_{t_i} |} \\
    S_i &= S - s_{t_{i-1}} - \cdots - s_{t_1} \\
    |s*|_i &= |s*| - |s_{t_{i-1}}| - \cdots - |s_{t_1}|
\end{aligned}
\end{equation}
where the $|\cdot|$ represent the operator to get the set of node labels. After $k$ steps, we get a set of skeleton candidates $ \{s_1, \cdots, s_k\}$.
% If $k>m$, we can extend $S$ by appending skeletons at distance of $d_0 + 1$.

Moreover, the input content such as relations and entities can also be taken into account.
In summary, the first priority for selecting examples is structural similarity, followed by the shared content.
%For example, in KoPL we extract the function's input arguments as the content. After we get a candidate pool $\{(l,q)| f(l)= s_1\} \cup \cdots \cup \{(l,q)| f(l)= s_k \}$ in the last step, we finally select the example pairs that share the most content with the target $l$. 

\subsection{Formal Language Generation}
\label{sec: flg}
In this task, the input of LLMs is the NLQ $q*$, so we search for example pairs $(q, l)$ from $\mathcal{S}$ where $q$s are most similar to $q*$.
%This task is for LLMs to directly translate a natural language question into the its correct logical forms, which is defined as the measure of LLMs' ability to generate structured semantics. The input here is the natural language question $q*$, so we should use the natural language question to search similar question examples from seed set $\mathcal{S} = \{ (l, q) \}$. 
%Therefore, we cannot use the structure similarity to find the best examples, because we cannot introduce $q*$'s structure information to the input. What we could the surface lexicon information

Similar to the previous task, we hope that the retrieved questions has a similar structure with $q*$.
Therefore, we utilize the BM25 algorithm to search in the seed set. As shown in Figure~\ref{fig:intro}, when constructing the BM25 searcher, we mask the entities and relations in the question to exclude their interference, hoping the searcher to pay more attention on conjunctions and prepositions that can potentially express the structure.

%Therefore, we intuitively utilize the structure prompt word such as ``and'' ``or'', which might indicate the concatenation structure in the question. Specifically, we use the BM25 algorithm to search similar questions with the same prompt words. As shown in Figure~\ref{fig:intro}, when constructing the BM25 searcher, we mask the entities and relations in the question to exclude their interference on the BM25 algorithm. This search method cannot guarantee the structure similarity, but it is a viable baseline method.

We did not adopt embedding-based searching algorithm such as BERT-Score because they can easily neglect conjunctions and prepositions, focusing on the semantics rather than structure. Besides, calculation speed is also in consideration.
%In fact, there are other ``better'' approaches, such as parsing the questions in the examples of $\mathcal{S}$ in advance to obtain their abstract syntax trees (ASTs), and then search for structurally similar questions based on ASTs. However, we argue that using AST in retrieval is leaking structural information in input to the LLMs, so here we only use the surface lexicon information of $q*$ to search for examples.

\subsubsection{Entity Linking}
\label{sec:el}
We found that the model often fails to generate the correct labels of entities and relations in the knowledge base. 
Therefore, we adopt several entity linking techniques to guide the model in generating correct labels. 

Before the generation, we retrieve some entities and relations that related to the question by matching the topic entity and finding relations with two hops, and then directly add them into the prompt.

After the generation, we check the generated LF and replace the inaccurately generated labels into correct labels by matching the most similar names in KB using BM25, similar to \citeauthor{DBLP:journals/corr/abs-2305-01750}. In this step, we also substitute the LLM-generated friendly names of entities into ids in KB, if needed.

\subsubsection{Chain-of-Thought Generation}
Due to the difficulty in directly generating LF, inspired by \citeauthor{DBLP:conf/emnlp/LiangWZWQL23}, we also adopt a multi-step chain-of-thought generation approach, which involves first generating the skeleton of LF and then filling in parameters for the complete LF. Details are in the appendix.

%Therefore, to generate correct and executable programs, we need to map the generated entities to the corresponding entries in the knowledge base. Taking the example of SPARQL based on Freebase, SPARQL statements use Freebase's \textit{mid} to represent entities, but the model struggles to generate the correct \textit{mid}. Similar to \citeauthor{DBLP:journals/corr/abs-2305-01750}, we have the model first generate the natural language names of the entities. Then, we use the BM25 algorithm to find the most similar entities in the entity repository. For concepts and relations in the knowledge base such as \textit{type.object.type} and \textit{rail.rail\_network}, although they are not represented by \textit{mid}, they are also often inaccurately generated. Thus, we use the BM25 algorithm to search for the most similar names in the knowledge base and replace them accordingly.
%Apart from correcting them after the generation, we find that other tricks such as adding entities and relations related to the question into the prompt also help improving the performance. More details and examples of this whole process including intermediate steps are provided in Appendix.

\section{Experiment Setup}
% 使用的模型，evaluation的方式，对每个dataset的parser的训练等
We introduce a range of popular language models that have been extensively studied in our experiments (\ref{sec: inv_mod}) as long as the semantic parsing models we use to evaluate the performance of the understanding task (\ref{sec: pef_eva}).

\subsection{Investigated Models}
\label{sec: inv_mod}
In order to investigate the impact of the model scale on its capacity, we select models of different sizes. 
%In this paper, we call small and medium size models (less than 100B) \textbf{Pretrained Language Models} (PLMs) and large size models (over 100B) \textbf{Large Language Models} (LLMs).

For medium size models ranging from 100M to 10B, we mainly consider two families of models. The first is auto-regressive models, exemplified by the GPT series. These models only use the decoder in training and employ a unidirectional ``predict the next word'' auto-regressive loss function for modeling. The second is represented by T5, a text-to-text model, which utilizes a bidirectional encoder and a unidirectional decoder to predict masked spans.
%This architecture and training approach strike a balance between the auto-regressive model of the GPT series and the masked language model of the BERT series. 
%T5 is trained by predicting corrupted spans in the text in the auto-regressive way. 
In the experiment, we use the instruction-tuned version FLAN-T5 series. The last is the open-source Llama-2 family, which is also modeled through an auto-regressive approach.
In particular, we select \textbf{GPT2-Large} (774M), \textbf{GPT2-XL} (1.5B)~\cite{radford2019language}, \textbf{GPT-J} (6B)~\cite{gpt-j}, \textbf{FLAN-T5-L} (770M), \textbf{FLAN-T5-XL} (3B), \textbf{FLAN-T5-XXL} (11B)~\cite{DBLP:journals/corr/abs-2210-11416}, \textbf{Llama-2-7B}, \textbf{Llama-2-13B}, \textbf{Llama-2-70B}~\cite{DBLP:journals/corr/abs-2307-09288}.

For large models over 100B, we first consider the instruction-tuned GPT 3.5 series, including the initial Davinci model \textbf{text-davinci-001} and the most powerful \textbf{text-davinci-003} (maybe 175B). We also investigate \textbf{GLM-130B}~\cite{DBLP:conf/iclr/ZengLDWL0YXZXTM23}, an open bilingual pretrained model without instruction-tuning and RLHF. We do not evaluate chat models like gpt-3.5-turbo since it is only considered the chat-optimized version of text-davinci-003, and under-performs davinci models in our pilot test. The code-pretrained model like CODEX is also not included because of it has closed access, and text-davinci-003, which has also been trained on code, can serve as a good substitute.

\iffalse
\begin{table}[]
    \centering
    \begin{tabular}{ccccc}
    \toprule
         Dataset & Train  & Dev  & Test  & Structure \\
    \midrule
          KQA Pro & 94376 & 11797 & 11797 & tree \\ 
          GrailQA & 44337 & 6763 & 13231 & graph \\
          Overnight & 8753 & 2189 & 2740 & tree \\
    \bottomrule
    \end{tabular}
    \caption{Dataset.}
    \label{tab:dataset}
\end{table}
\fi
\subsection{Evaluation Models}
The evaluation methods is mentioned above in~\ref{sec: pef_eva}, 
%we evaluate the quality of the generated natural language question by comparing the semantic parser's performance trained by the model-generated questions to the parser trained on human-labelled data. 
In practice, different semantic parsers are chosen for the evaluation of different formal languages and datasets.

%The statistics of the datasets and the split we employed is shown in Table~\ref{tab:dataset}. 
For KoPL and KQA Pro dataset, we use the original baseline (BART-base) provided  KQA Pro~\cite{DBLP:conf/acl/CaoSPNX0LHZ22}. For Lambda DCS and Overnight dataset, we train a bidirectional LSTM with dual learning algorithm described by \citeauthor{cao-etal-2019-semantic}. Finally, for SPARQL and GrailQA, we tried two baseline models. One is also a simple sequence-to-sequence BART-base generation model without explicit entity linking modules. 
%The only trick is to add the corresponding entity and relation names in KB to the vocabulary of the parser, which is similar to the practice described in the GrailQA's original baseline~\cite{DBLP:conf/www/GuKVSLY021}. 
The other baseline is a rank-and-generate (RnG) pipeline with an entity linking module described in \citeauthor{DBLP:conf/acl/YeYHZX22}, which 
employs a ranker to retrieve related logical forms that share similar entities and relations.
%first employ a ranker to select related logical forms from a candidate pool obtained by searching over the graph, then generate the final output conditioned on both the input question and the top-ranked logical forms. 
The implementation detail of parsers and training hyper-parameters used in the work can be found in Appendix.

\section{Results and Analyses}
\label{sec:ana_con}
We first present the main result of the formal language understanding and generation in Table~\ref{tab:1}.

In the left blue section of understanding task, the figures are the absolute performance of the evaluation parser trained on training sets that generated by different models. The retrieved examples of the input prompt of ICL is 3 for all models in the understanding task. 

The right green section presents the semantic parsing result of the models, where the retrieved examples are as many as the input context can take so as to improve the result. To cut down computation overehead, the test sets are randomly sampled subsets of 300, 120, 240 examples from the test sets of KQA Pro, GrailQA, and Overnight, respectively. The parsing performance of KoPL and Lambda DCS are measured by answers' accuracy, and the SPARQL performance are measure by answers' F1 score. Note that the human's performance is not applicable here, but we can compare it to the baseline results of understanding task. Also, we only test the model over 1B because the small models perform poorly with meaningless results.

Then we present the conclusions and findings by analyzing them along with other ablation experiments. More detailed results for some dataset can be found in Appendix. 

\begin{table*}[ht]
\centering
\scalebox{0.9}
%\resizebox{\textwidth}{!}
{
\begin{tabular}{l|cccc|ccc}
\toprule
\multirow{2}{*}{\textbf{Model}} & \multicolumn{4}{c}{\cellcolor{'shallow1'}\textbf{Understanding}} & \multicolumn{3}{|c}{\cellcolor{'shallow2'}\textbf{Generation}} \\
\cmidrule(l){2-8}
&\cellcolor{'deep1'} KoPL &\cellcolor{'deep1'}SPARQL & \cellcolor{'deep1'}SPARQL$_\ddag$ &  \cellcolor{'deep1'}Lambda DCS & \cellcolor{'deep2'}KoPL & \cellcolor{'deep2'}SPARQL & \cellcolor{'deep2'}Lambda DCS \\

%  metric   & acc & acc & acc & acc & acc & f1  \\
\midrule
\cellcolor{'wit'}GPT2-L (774M) & \cellcolor{'shallow1'}76.0 &  \cellcolor{'shallow1'}70.8 & \cellcolor{'shallow1'}10.8 & \cellcolor{'shallow1'}39.1 & \cellcolor{'shallow2'}--- & \cellcolor{'shallow2'}--- & \cellcolor{'shallow2'}---\\
\cellcolor{'gry'}GPT2-XL (1.5B) & \cellcolor{'deep1'}83.3 &  \cellcolor{'deep1'}71.1 & \cellcolor{'deep1'}14.4 & \cellcolor{'deep1'}42.3 & \cellcolor{'deep2'}--- & \cellcolor{'deep2'}--- & \cellcolor{'deep2'}---\\
\cellcolor{'wit'}GPT-J (6B) & \cellcolor{'shallow1'}84.2 & \cellcolor{'shallow1'}72.2 &  \cellcolor{'shallow1'}16.7 & \cellcolor{'shallow1'}74.4 & \cellcolor{'shallow2'}4.3 & \cellcolor{'shallow2'}1.7 & \cellcolor{'shallow2'}0.0\\
\cellcolor{'gry'}FLAN-T5-L (770M) & \cellcolor{'deep1'}48.6 & \cellcolor{'deep1'}71.6 & \cellcolor{'deep1'}6.8 & \cellcolor{'deep1'}27.5 & \cellcolor{'deep2'}--- & \cellcolor{'deep2'}--- &\cellcolor{'deep2'}---\\
\cellcolor{'wit'}FLAN-T5-XL (3B) & \cellcolor{'shallow1'}26.6 & \cellcolor{'shallow1'}70.7& \cellcolor{'shallow1'}7.1 &  \cellcolor{'shallow1'}17.0 & \cellcolor{'shallow2'}--- & \cellcolor{'shallow2'}--- & \cellcolor{'shallow2'}---\\
\cellcolor{'gry'}FLAN-T5-XXL (11B) & \cellcolor{'deep1'}12.7 & \cellcolor{'deep1'}68.1 & \cellcolor{'deep1'}7.0 & \cellcolor{'deep1'}12.4 & \cellcolor{'deep2'}2.7 & \cellcolor{'deep2'}0.0 & \cellcolor{'deep2'}0.0\\

\cellcolor{'wit'}Llama-2-7B & \cellcolor{'shallow1'}83.8 & \cellcolor{'shallow1'}71.2 & \cellcolor{'shallow1'}16.6 & \cellcolor{'shallow1'}73.2 & \cellcolor{'shallow2'}4.6 & \cellcolor{'shallow2'}1.7 & \cellcolor{'shallow2'}0.0\\

\cellcolor{'gry'}Llama-2-13B & \cellcolor{'deep1'}85.2 & \cellcolor{'deep1'}71.9 & \cellcolor{'deep1'}17.1 & \cellcolor{'deep1'}74.6 & \cellcolor{'deep2'}10.0 & \cellcolor{'deep2'}2.5 & \cellcolor{'deep2'}0.0\\

\cellcolor{'wit'}Llama-2-70B & \cellcolor{'shallow1'}85.8 & \cellcolor{'shallow1'}72.6 &
\cellcolor{'shallow1'}18.5 & \cellcolor{'shallow1'}75.3 & \cellcolor{'shallow2'}11.3 & \cellcolor{'shallow2'}4.2 & \cellcolor{'shallow2'}3.3\\
\cellcolor{'gry'}GLM-130B & \cellcolor{'deep1'}86.2 & \cellcolor{'deep1'}73.6 & \cellcolor{'deep1'}19.2 & \cellcolor{'deep1'}77.0 & \cellcolor{'deep2'}22.3 & \cellcolor{'deep2'}5.8 & \cellcolor{'deep2'}3.8\\
\cellcolor{'wit'}Text-Davinci-001 & \cellcolor{'shallow1'}85.6 & \cellcolor{'shallow1'}71.4 & \cellcolor{'shallow1'}18.7 & \cellcolor{'shallow1'}75.2 & \cellcolor{'shallow2'}16.0 & \cellcolor{'shallow2'}2.7 & \cellcolor{'shallow2'}1.7\\
\cellcolor{'gry'}Text-Davinci-003 & \cellcolor{'deep1'}88.1 & \cellcolor{'deep1'}73.8 & \cellcolor{'deep1'}21.7 & \cellcolor{'deep1'}79.0 & \cellcolor{'deep2'}41.6 & \cellcolor{'deep2'}22.5 & \cellcolor{'deep2'}10.0 \\
\midrule
\cellcolor{'wit'}Human & \cellcolor{'shallow1'}90.6 & \cellcolor{'shallow1'}74.7 & \cellcolor{'shallow1'}28.1 & \cellcolor{'shallow1'}95.2 & \cellcolor{'shallow2'}---& \cellcolor{'shallow2'}--- & \cellcolor{'shallow2'}---\\

\bottomrule
\end{tabular}}
\caption{The main results of formal language understanding and generation. $\ddag$ means that these column is evaluated by a simple sequence-to-sequence Bart-base parser without an entity linking module. The --- in the table means the result is trivial because of the low performance or it is not applicable.}
\label{tab:1}
\end{table*}

\subsection{Formal Language Understanding Result Analysis}
As shown in Table~\ref{tab:1}, we can see that (1) All language models demonstrate a certain degree of understanding of formal languages, as evidenced by their ability to generate new training data to train a non-trivial parser. (2) In general, larger models tend to perform better in understanding structured semantics. (3) LLMs are sensitive to formal languages. For example, their performance on KOPL and SPARQL is noticeably closer to human-level. This might be attributed to the pre-training data. (4) As for the parser for text quality evaluation, the RnG parser can virtually eliminate gaps in generated data quality, reflecting the importance of entity linking module. (5) Meanwhile, it is noteworthy that we do not observe significant differences between models that are instruction tuned and those that are not. The model size evidently has a more pronounced impact. 

Most interestingly, 
%We observed strong sensitivity of understanding performance to both LLMs and formal languages. The FLAN-t5 series are particularly unusual that, as the model size increases, the performance deteriorates. In the appendix, we present some error analysis from FLAN-T5-XXL, whose generated results are almost unintelligible.
We observe some peculiar characteristics in the FLAN-T5 series. Not only do they perform significant worse compared to other models of similar scale, but more unusually, the performance deteriorates as the model size increases. In the appendix, we present some error analysis from FLAN-T5-XXL, whose generated results are almost unintelligible.

\begin{figure}[ht]
    \centering
    \includegraphics[width=.5\textwidth]{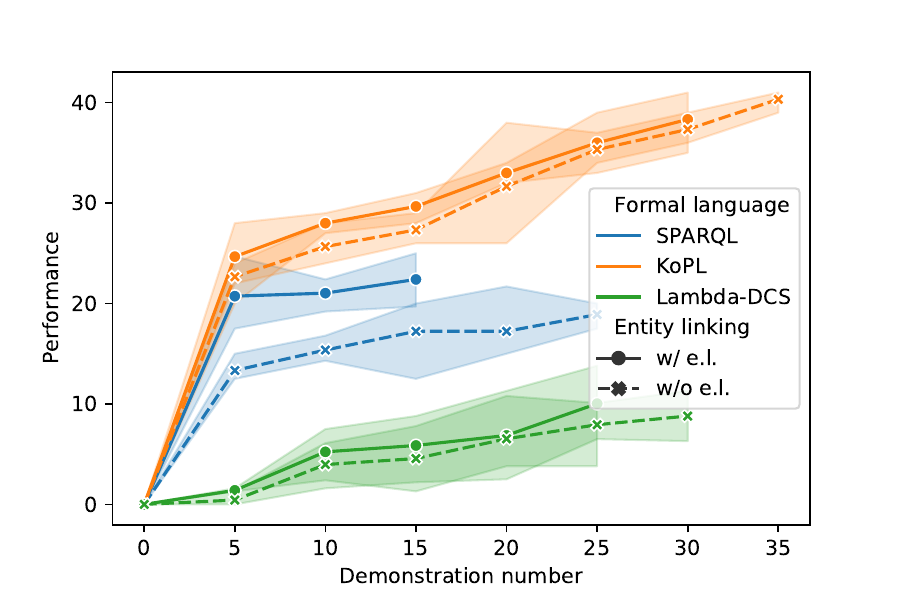}
    \caption{\label{fig:gen_demonum} Formal language generation performance of Text-Davinci-003 with various numbers of demonstration examples. The entity linking tag means whether to use entity linking to detect the entities in input and add their 2-hop-related entity and relation names to the input. Note that the difference of maximum demonstration number between formal languages is because the context length of LLM. Each data point takes 3 runs and details in appendix~\ref{sec: flg_detail}. }
\end{figure}

\subsection{Formal Language Geration Result Analysis}
From the right section of Table~\ref{tab:1}, we can see that the generation ability of language models is far worse than their understanding ability. Compared to the left section, even the most powerful model directly generating logical forms can only achieve 15\% to 50\% accuracy to the parser trained by its generated data. Therefore, we believe it is safe to reach to the conclusion that, to improve performance on knowledge based question answering, it is much more easier to generate new data for training small parser like Bart model than directly using LLMs to generate if we only wish to prompt without touching the parameters.

To improve the performance of direct semantic parsing, two approaches are viable in the experiment. The first is increasing the examples of ICL and the second is to detect the entities mentioned in the input question, and include their 2-hop-related entity and relation names from the knowledge base into the prompt (as mentioned in Section\ref{sec:el}). To compare the impact of these two strategies on the performance, we conduct a series of experiment on Text-Davinci-003. As shown in Figure~\ref{fig:gen_demonum}, (1) Both strategies can contribute to the performance. (2) The performance on KoPL notably improves with the increase of examples. However, for SPARQL and Lambda DCS, the effect of this strategy is limited. (3) On the other hand, incorporating entity and relation names in the prompt significantly enhances the results for SPARQL. (4) In all settings, model performs best on KoPL and worst on Lambda DCS, and SPARQL in between. 

Empirically, We figure the possible explanations for these phenomena lie in the difference between formal languages. As the example show in Figure~\ref{fig:intro_b}, KoPL is the most similar to natural language. The identifiers are easy for human to understand, and the order of functions correspond to the ``chain-of-thought'' reasoning process. While both SPARQL and Lambda DCS are more formalized and contain lots of identifiers that do not make sense in natural language. This might explain why model performs best on KoPL, and most benefits from the increasing of examples. Furthermore, we note that the grammar of SPARQL is simpler and lacks of variations, where the SPARQL queries in the GrailQA dataset almost follow the same pattern. But the bottle-neck for writing SPARQL is to generate the correct entity or relation names in Freebase. This explains why model performs better on SPARQL than Lambda DCS, and why adding entities to prompt improves the most for SPARQL.

\subsection{Zero-shot Understanding}
We are also very interested in whether the LLMs truly understand the logical forms or they merely are good at imitating the carefully selected examples we provided? To figure it out, we conduct an ablation experiment where input for the QG task is replaced with the description of the formal language. This experiment is only conducted on KoPL since it is well modularized and the function of the operations can be concisely explained. The input description consists of the one-sentence descriptions of each operation function in KoPL, optionally accompanied by several fixed simple examples. To reduce the cost, we only use a subset that contain the first 20,000 examples of KoPL ( the same in next experiment in Section \ref{sec:dif_ratio}) and only probe the GPT series.

\begin{table}[ht]
    \centering
    \resizebox{0.45\textwidth}{!}{
    \begin{tabular}{l|cc}
    \toprule
        Model & KoPL$_{1\%seed}$ & KoPL$_{zero-shot}$ \\
    \midrule
        GPT-J (6B) & 43.3 & 11.9 \\
        GLM-130B & 76.4 & 46.0 \\
        Text-Davinci-001 & 76.8 & 44.6\\
        Text-Davinci-003 & 80.0 & 62.7\\
    \midrule
        Human & 84.6 & 84.6 \\
    \bottomrule
    \end{tabular}}
    \caption{Formal language understanding results for the low-resource seed set setting and the zero-shot setting.}
    \label{tab:2}
\end{table}

As shown in the Table~\ref{tab:2}, it can be observed that the carefully designed retrieval strategy in our baseline method indeed significantly contributes to generating high-quality natural language questions. However, at the same time, the model itself exhibits a certain degree of understanding ability when examples are lacking, where Text-Davinci-003 demonstrates a 25.8\% performance drop.

\subsection{Different Seed Set Ratio}
\label{sec:dif_ratio}
%Since our initial goal is to explore LLMs' ability, and considering the possible difficulty of these tasks, we test LLMs' performance under the condition that the seed set $\mathcal{S}$ for generation is the training set, which maximize the probability of retrieving similar demonstration examples. 
The main result in our experiment are generated with the whole training set as the seed set. However, considering the practical limitations in obtaining a large amount of high-quality manually annotated data in real scenarios, we investigate the model's ability to generate new data with only a small amount of labeled data as seeds.

This experiment is also conducted on KQA Pro since it is the largest and most diverse dataset. We randomly sample 1\% of training set as seeds. The result in Table~\ref{tab:2} indicate that although there is a decrease in the quality of generated questions, the performance degradation of the model is acceptable, given the great reduction in seed number.

\begin{table}[ht]
    \centering
\scalebox{0.9}
%\scalebox{0.9}
{
    \begin{tabular}{l|cc}
    \toprule
    \multirow{2}{*}{Model} & \multicolumn{2}{c}{Understanding}  \\
    \cmidrule(l){2-3}
     & SPARQL & Lambda DCS \\
    \midrule
        GPT-J (6B) & 71.9 & 62.4 \\
        GLM-130B & 76.3 & 64.8 \\
        Text-Davinci-001 & 74.4 & 61.6 \\
        Text-Davinci-003 & 80.2 & 69.7 \\
    \midrule
        Human & 82.7 & 76.1 \\
    \toprule
    \multirow{2}{*}{Model}& \multicolumn{2}{c}{Generation} \\
    \cmidrule(l){2-3}
     & SPARQL & Lambda DCS \\
    \midrule
        Text-Davinci-003 & 14.2 & 4.2\\
    \bottomrule
    \end{tabular}}
    \caption{Formal language understanding and generation results for the one-dataset setting.}
    \label{tab:3}
\end{table}

\subsection{All Formal Languages on One Dataset}
Since different datasets are constructed on different knowledge bases, in order to compare whether the three logic forms can arrive at the similar conclusions on identical data as previously observed, we conduct a experiment testing the three formal languages on the same dataset.

This experiment is also conducted on KQA Pro for convenience, because it already contains KoPL and SPARQL, and the parser for evaluation also switches to BART-base, the same with KQA Pro. And we follows \citeauthor{DBLP:conf/emnlp/NieCSST0LZ22} to translate KoPL into Lambda DCS. From results in Table~\ref{tab:3}, overall the results are consistent to the main result in Table~\ref{tab:1}. But the performance of generation drops a bit, because for SPARQL, the entity and relation binding process are skipped in this experiment. 

%that it is of better language understanding for a intelligence system to comprehend the deep structure, namely the derivation trees of sentence, which is explicitly represented in the human-designed formal language.

\section{Conclusion}
%In this work, we leverage the formal language to probe the deep structure understanding of natural language in LLMs. Our observations suggest that there still exists a gap between LLMs and human. 
In this work, we evaluate the proficiency of different LLMs in understanding and generating different formal languages. Our observations suggest that the ability of LLMs to generate structured semantics is notably inferior to their ability to understand it.
More importantly, 
LLMs demonstrate the sensitivity to different formal languages. Aligning with our intuition, we discover that the choice of formal language and knowledge base can exert significant influence on models' performance. 

In our experiment, models performing on KoPL yields the best results on nearly all experiments. We believe it is because KoPL employs expressions that are more similar to natural language while preserving the structure and modularity. 
%Importantly, its expression of the reasoning process closely aligns with human. 
However, SPARQL and Lambda DCS face challenges in grounding entities to the knowledge base for their level of formalization is too high. As a result, KoPL proves to be the most LLMs-friendly 
%for semantic parsing task 
among the formal languages that we investigate in this work.
%对于KoPL来说，它的总体形式化程度较低，最接近自然语言。其次它的模块化较好，每个函数的功能都能用简单的自然语言表述，而且函数名和输入的参数都是用自然语言表达的。最重要的是，它所表达的推理过程也非常接近于CoT。因此几乎所有实验在KoPL上的效果都是最好的，即使在使用zero-shot 或 较低比例的种子集的情况下，都仍能有较好的表现。在语义解析的任务上，它也是最LLMs-friendly的，因为可以直接通过只增加例子数量提高效果

%SPARQL's advantage is the consistent pattern observed in the majority of SPARQL queries within the dataset, making the formatting and syntax relatively simple. However, the challenge lies in the fact that the entity identifiers do semantic information, and the relationship names in the knowledge base are often complex. Therefore, the bottleneck lies in entity linking. On the other hand, Lambda DCS exhibits a more varied grammar and faces the challenge of grounding entity and relation names to the knowledge base as well. As a result, we observe that model usually not good at dealing with Lambda DCS.
%SPARQL的优势在于数据集中大多的sparql查询都遵循同样的pattern，因此写起来格式和语法不是问题。但困难之处在于，SPARQL query的实体使用的是没有语义信息的mid，而关系名称在知识库中也大多相对复杂，因此其主要瓶颈在于实体对齐，而非大模型本身的生成能力。

%However, due to the relatively simple nature of the overnight dataset we use, with only a few entities involved, the difficulty in composing Lambda DCS queries was not prominently reflected in the experiments.
% 而Lambda DCS的语法相对更加多变，同时也存在和SPARQL一样的需要将实体关系名称对齐到知识库的问题，

In general, we want to point out that the formal language plays an important role in enhancing the power of LLMs. A formal language can be used as a medium between LLMs and the knowledge base, so that LLMs can use the knowledge base as a tool to enhance the performance of QA and reasoning tasks. On the other hand, the selection of a more model-friendly formal language, one that closely resembles the natural language in which models excel, should be prioritized.

\section*{Limitations}
In this work, we do not systematically study the code-pretrained models. The main reason is that there are no 100B version in most code model series for comparison with non-code models. Additionally, OpenAI's CODEX, which was previously available, has been discontinued, and its functionality can be replaced by text-davinci-003. Considering the systematic nature of model selection, we did not choose code models. 
Another limitation is we only study the LLMs' proficiency in formal languages as a whole. Later we will consider designing tasks such as completion to conduct more detailed research.
%This is mainly \textcolor{red}{because the most powerful model \textbf{CODEX} is not available}, and other smaller code models are not systematically comparable to the non-code models.

\section{Acknowledgments}
We would like to thank all the anonymous reviewers for their careful work and thoughtful suggestions. This work is supported by a grant from the Institute for Guo Qiang, Tsinghua University (2019GQB0003) and Tsinghua University Initiative Scientific Research Program. This work is also supported by Huawei Cloud Computing Technologies Co., Ltd. 

\bibliography{anthology, custom}

\appendix

\section{Details of Probing Process}
\label{sec:a1}
In this section we present the details of the probing processes of both probing sub-tasks.

\begin{table*}[ht]
    \centering
    \begin{tabular}{p{\linewidth}}
    \toprule
        \cellcolor{'shallow1'}\textbf{\textsc{Natural Language Question}} : What is the name of the actor that was born in 1956-04-19 ? \\
    \midrule
        \cellcolor{'shallow1'}\textbf{\textsc{KoPL Program}}: FindAll().FilterDate(date of birth, 1956-04-19, =).FilterConcept(human).Find(actor).Relate(occupation, backward).FilterConcept(human).And().What()\\
    \midrule
        \cellcolor{'shallow1'}\textbf{\textsc{KoPL Skeleton}}: FindAll.FilterDate.FilterConcept.Find.Relate.FilterConcept.And.What\\
    \midrule
        \cellcolor{'shallow1'}\textbf{\textsc{Prompt}}: According to the given logic form kopl, generate the corresponding natural language question. For examples, \\\cellcolor{'shallow1'}FindAll()FilterDate(date of birth, 1989-04-06, =)FilterConcept(human)Find(United States of America)Relate(country of citizenship, backward)FilterConcept(human)And()What() is verbalized as: Which human was born 1989-04-06 and is a citizen of the United States of America? [SEP] \\ \cellcolor{'shallow1'}FindAll()FilterDate(date of birth, 1977-03-10, =)FilterConcept(human)Find(association football)Relate(sport, backward)FilterConcept(human)And()What() is verbalized as: Which human has the date of birth 1977-03-10 and is related to the sport association football? [SEP] \\\cellcolor{'shallow1'}FindAll()FilterDate(date of birth, 1956-04-19, =)FilterConcept(human)Find(actor)Relate(occupation, backward)FilterConcept(human)And()What() is verbalized as: What is the name of the actor that was born in 1956-04-19? [SEP] \\\cellcolor{'shallow1'}  FindAll()FilterStr(TOID, 4000000074573917)FilterConcept(town)FindAll()FilterStr(OS grid reference, SP8778)FilterConcept(town)And()What() is verbalized as: \\
    \bottomrule
    \end{tabular}
    \caption{Serialized examples of the KoPL and its corresponding skeletons, and final input prompt.}
    \label{tab:sk1}
\end{table*}
\subsection{Formal Language Understanding}
In this task, we search the seed set for demonstration examples based on the structure of the input logical form. As stated in the main submission, we first transform the logical forms into corresponding skeletons.

The skeleton of KoPL is the tree formed by removing the functions' inputs, and we serialize it with post-order traversal. The Lambda DCS program is similar, it is a bracket tree and its skeleton is also also formed by only keeping identifiers. The SPARQL program depicts a graph by some triples, and the algorithm for finding graphs with the same structure is complex, so we use the SPAQRL's corresponding S-expression, which is also a bracket tree structure. The serialized examples of the logical form skeleton is illustrated in Table~\ref{tab:sk1}, respectively.

\begin{table*}[ht]
    \centering
    \begin{tabular}{p{\linewidth}}
    \toprule
        \cellcolor{'shallow1'}\textbf{\textsc{Natural Language Question}}: What format does the station which broadcasts mojo in the morning use?\\
    \midrule
        \cellcolor{'shallow1'}\textbf{\textsc{SPARQL Program}}: SELECT (?x0 AS ?value) WHERE \{ SELECT DISTINCT ?x0 WHERE \{ ?x0 :type.object.type :broadcast.radio\_format . ?x1 :type.object.type :broadcast.radio\_station . VALUES ?x2 \{ :m.010fcxr0 \} ?x1 :broadcast.radio\_station.format ?x0 . ?x1 :broadcast.broadcast.content ?x2 . FILTER ( ?x0 != ?x1 \&\& ?x0 != ?x2 \&\& ?x1 != ?x2 ) \} \} \\
    \midrule
        \cellcolor{'shallow1'}\textbf{\textsc{S-expression}}: (AND broadcast.radio\_format (JOIN (R broadcast.radio\_station.format) (JOIN broadcast.broadcast.content m.010fcxr0))) \\
    \midrule
        \cellcolor{'shallow1'}\textbf{\textsc{SPARQL Skeleton}}: (AND [V0] (JOIN (R [V1]) (JOIN [V2] [E0])))\\
    \midrule
        \cellcolor{'shallow1'}\textbf{\textsc{Prompt}}: According to the given logic form sparql, generate the corresponding natural language question. For examples, \\\cellcolor{'shallow1'} SELECT (?x0 AS ?value) WHERE \{ SELECT DISTINCT ?x0 WHERE \{ ?x0 :type.object.type :broadcast.producer . ?x1 :type.object.type :broadcast.content . VALUES ?x2 \{ :latino \} ?x1 :broadcast.content.producer ?x0 . ?x1 :broadcast.content.genre ?x2 . FILTER ( ?x0 != ?x1 \&\& ?x0 != ?x2 \&\& ?x1 != ?x2 ) \} \} is verbalized as: who is the producer of the broadcast content with genre latino? [SEP]\\\cellcolor{'shallow1'} SELECT (?x0 AS ?value) WHERE \{ SELECT DISTINCT ?x0 WHERE \{ ?x0 :type.object.type :broadcast.producer . ?x1 :type.object.type :broadcast.content . VALUES ?x2 \{ :90's \} ?x1 :broadcast.content.producer ?x0 . ?x1 :broadcast.content.genre ?x2 . FILTER ( ?x0 != ?x1 \&\& ?x0 != ?x2 \&\& ?x1 != ?x2 ) \} \} is verbalized as: who produces 90's genre broadcast content? [SEP]\\\cellcolor{'shallow1'} SELECT (?x0 AS ?value) WHERE \{ SELECT DISTINCT ?x0 WHERE \{ ?x0 :type.object.type :broadcast.producer . ?x1 :type.object.type :broadcast.content . VALUES ?x2 \{ :audio podcast \} ?x1 :broadcast.content.producer ?x0 . ?x1 :broadcast.content.genre ?x2 . FILTER ( ?x0 != ?x1 \&\& ?x0 != ?x2 \&\& ?x1 != ?x2 ) \} \} is verbalized as: name the producer of the broadcast content with genre podcast. [SEP]\\\cellcolor{'shallow1'} SELECT (?x0 AS ?value) WHERE \{ SELECT DISTINCT ?x0 WHERE \{ ?x0 :type.object.type :broadcast.radio\_format . ?x1 :type.object.type :broadcast.radio\_station . VALUES ?x2 \{ :mojo \} ?x1 :broadcast.radio\_station.format ?x0 . ?x1 :broadcast.broadcast.content ?x2 . FILTER ( ?x0 != ?x1 \&\& ?x0 != ?x2 \&\& ?x1 != ?x2 ) \} \} is verbalized as: \\
    \bottomrule
    \end{tabular}
    \caption{Serialized examples of the SPARQL and its corresponding skeletons, and final input prompt. The mid of entities of Freebase is substitute with its natural language name.}
    \label{tab:sk2}
\end{table*}

\begin{table*}[ht]
    \centering
    \begin{tabular}{p{\linewidth}}
    \toprule
        \cellcolor{'shallow1'}\textbf{\textsc{Natural Language Question}}: What players made less than three assists over a season?\\
    \midrule
       \cellcolor{'shallow1'}\textbf{\textsc{Lambda DCS Program}}: ( call SW.listValue ( call SW.getProperty ( ( lambda s ( call SW.filter ( var s ) ( call SW.ensureNumericProperty ( string num\_assists ) ) ( string $<$ ) ( call SW.ensureNumericEntity ( number 3 assist ) ) ) ) ( call SW.domain ( string player ) ) ) ( string player ) ) ) \\
    \midrule
        \cellcolor{'shallow1'}\textbf{\textsc{Lambda DCS Skeleton}}: ( call SW.listValue ( call SW.getProperty ( ( lambda ( call SW.filter ( var ) ( call SW.ensureNumericProperty ( string ) ) ( string ) ( call SW.ensureNumericEntity ( number ) ) ) ) ( call SW.domain ( string ) ) ) ( string ) ) )\\
    \midrule
        \cellcolor{'shallow1'}\textbf{\textsc{Prompt}}: According to the given logic form lambdaDCS, generate the corresponding natural language question. For examples, \\\cellcolor{'shallow1'} ( call SW.listValue ( call SW.getProperty ( ( lambda s ( call SW.filter ( var s ) ( call SW.ensureNumericProperty ( string num\_assists ) ) ( string $<$ ) ( call SW.ensureNumericEntity ( number 3 assist ) ) ) ) ( call SW.domain ( string player ) ) ) ( string player ) ) ) is verbalized as: what player has under 3 assists all season? [SEP]\\\cellcolor{'shallow1'} ( call SW.listValue ( call SW.getProperty ( ( lambda s ( call SW.filter ( var s ) ( call SW.ensureNumericProperty ( string num\_assists ) ) ( string $<$ ) ( call SW.ensureNumericEntity ( number 3 assist ) ) ) ) ( call SW.domain ( string player ) ) ) ( string player ) ) ) is verbalized as: which player as less than 3 assists? [SEP]\\\cellcolor{'shallow1'} ( call SW.listValue ( call SW.getProperty ( ( lambda s ( call SW.filter ( var s ) ( call SW.ensureNumericProperty ( string num\_assists ) ) ( string $<$ ) ( call SW.ensureNumericEntity ( number 3 assist ) ) ) ) ( call SW.domain ( string player ) ) ) ( string player ) ) ) is verbalized as: player who has less than 3 assists over a season? [SEP]\\\cellcolor{'shallow1'} ( call SW.listValue ( call SW.getProperty ( ( lambda s ( call SW.filter ( var s ) ( call SW.ensureNumericProperty ( string num\_assists ) ) ( string $<$ ) ( call SW.ensureNumericEntity ( number 3 assist ) ) ) ) ( call SW.domain ( string player ) ) ) ( string player ) ) ) is verbalized as: \\
    \bottomrule
    \end{tabular}
    \caption{Serialized examples of the Lambda DCS and its corresponding skeletons, and final input prompt.}
    \label{tab:sk3}
\end{table*}

\subsection{Formal Language Generation}
In this task, we search the seed set for demonstration examples only based on the input natural language question. As mentioned in the main paper Section~\ref{sec: flg}, we mask the entities and relations in the question to get the NLQ skeleton. And then the prompt is constructed in the chain-of-thought manner - first generating the skeleton, then adding arguments. We take KoPL as an example, where the demonstration number equals 3. The the input question, skeleton and final prompt is illustrated in Table~\ref{tab:sk4}. This method works the same for other two formal languages, so we will not continue to show examples here.

But the chain-of-thought prompt does not always work better. When evaluating Llama-2 models, we observed that generating directly from NLQ to complete LF often gets better performance.

\begin{table*}[ht]
    \centering
    \begin{tabular}{p{\linewidth}}
    \toprule
         \cellcolor{'shallow2'}\textbf{\textsc{Natural Language Question}} : Which cost less? Batman Begins
released in Italy or Tootsie. \\
    \midrule
        \cellcolor{'shallow2'}\textbf{\textsc{NLQ Skeleton}}: Which cost less? [E0] released in [E1] or [E2]. \\
    \midrule
        \cellcolor{'shallow2'}\textbf{\textsc{Prompt}}: According to the given natural language question, generate the corresponding logic form in kopl. For examples, \\\cellcolor{'shallow2'} When did the state with the motto of Dio, Patria e liberta have an inflation rate of 6 percentage? is parsed into: 
        \\\cellcolor{'shallow2'}Functions: Find [func] Relate [func] Find [func] And [func] Relate [func] FilterConcept
        \\\cellcolor{'shallow2'}Adding arguments: Find [arg] Walt Disney Pictures [func] Relate [arg] production company [arg] backward [func] Find [arg] Pocahontas [func] And [func] Relate [arg] film crew member [arg] forward [func] FilterConcept [arg] human [func] QueryAttrQualifier [arg] Twitter username [arg] TimAnimation [arg] number of subscribers [SEP]\\\cellcolor{'shallow2'} Did a person, who received s Primetime Emmy Award for Outstanding Guest Actress in a Comedy Series in 2005, die before 2017 ? is parsed into:
        \\\cellcolor{'shallow2'}Functions: Find [func] Relate [func] QFilterYear [func] FilterConcept [func] QueryAttr [func] VerifyYear
        \\\cellcolor{'shallow2'}Adding arguments: Find [arg] Primetime Emmy Award for Outstanding Guest Actress in a Comedy Series [func] Relate [arg] winner [arg] forward [func] QFilterYear [arg] point in time [arg] 2005 [arg] = [func] FilterConcept [arg] human [func] QueryAttr [arg] date of death [func] VerifyYear [arg] 2017 [arg] < [SEP]\\\cellcolor{'shallow2'} How many conservatories focus on art form s from Mexico ? is parsed into: 
        \\\cellcolor{'shallow2'}Functions: Find [func] Relate [func] FilterConcept [func] Relate [func] FilterConcept [func] Count
        \\\cellcolor{'shallow2'}Adding Arguments: Find [arg] Mexico [func] Relate [arg] country [arg] backward [func] FilterConcept [arg] art form [func] Relate [arg] field of work [arg] backward [func] FilterConcept [arg] conservatory [func] Count [SEP]\\\cellcolor{'shallow2'} Which cost less? Batman Begins
released in Italy or Tootsie? is parsed into: \\
    \bottomrule
    \end{tabular}
    \caption{An example in the formal language generation task, including the input natural language question, the correpsonding skeleton, and the final prompt.}
    \label{tab:sk4}
\end{table*}

\subsection{Zero-shot Understanding}
The prompt used in the experiment of zero-shot understanding is shown in Table~\ref{tab:zsu}
\begin{table*}[ht]
    \centering
    \begin{tabular}{p{\linewidth}}
    \toprule
        \cellcolor{'shallow1'}\textbf{\textsc{Zero-Shot Prompt}}: Introduction for the formal language KOPL is as followed.
\cellcolor{'shallow1'}KOPL is a query language for knowledge-based question answering. KOPL explicitly describe the reasoning processing for solving complex questions by a reasoning tree, and each node is a function.
The function library is as followed: \\
\cellcolor{'shallow1'}1. Findall(): Return all entities in KB. \\
\cellcolor{'shallow1'}2. Find(): Return all entities with the given name. \\
\cellcolor{'shallow1'}3. FilterConcept(): Find those belonging to the given concept. \\
\cellcolor{'shallow1'}4. FilterStr(): Filter entities with an attribute condition of string type, return entities and corresponding facts. \\
\cellcolor{'shallow1'}5. FilterNum(): Similar to FilterStr, but atrribute type is number. \\
\cellcolor{'shallow1'}6. FilterYear(): Similar to FilterStr, but attribute type is year. \\
\cellcolor{'shallow1'}7. FilterDate(): Similar to FilterStr, but attribute type is date. \\
\cellcolor{'shallow1'}8. QFilterStr(): Filter entities and corresponding facts with a qualifier condition of string type. \\
\cellcolor{'shallow1'}9. QFilterNum(): Similar to QFilterStr, but qualifier type is number. \\
\cellcolor{'shallow1'}10. QFilterYear(): Similar to QFilterStr, but qualifier type is year. \\
\cellcolor{'shallow1'}11. QFilterDate(): Similar to QFilterStr, but qualifier type is date. \\
\cellcolor{'shallow1'}12. Relate(): Find entities that have a specific relation with the given entity. \\
\cellcolor{'shallow1'}13. And(): Return the intersection of two entity sets. \\
\cellcolor{'shallow1'}14. Or(): Return the union of two entity sets. \\
\cellcolor{'shallow1'}15. QueryName(): Return the entity name. \\
\cellcolor{'shallow1'}16. Count(): Return the number of entities. \\
\cellcolor{'shallow1'}17. QueryAttr(): Return the attribute value of the entity. \\
\cellcolor{'shallow1'}18. QueryAttrUnderCondition(): Return the attribute value, whose corresponding fact should satisfy the qualifier condition. \\
\cellcolor{'shallow1'}19. QueryRelation(): Return the relation between two entities. \\
\cellcolor{'shallow1'}20. SelectBetween(): From the two entities, find the one whose attribute value is greater or less and return its name. \\
\cellcolor{'shallow1'}21. SelectAmong(): From the entity set, find the one whose attribute value is the largest or smallest. \\
\cellcolor{'shallow1'}22. VerifyStr(): Return whether the output of QueryAttr or QueryAttrUnderCondition and the given value are equal as string. \\
\cellcolor{'shallow1'}23. VerifyNum(): Return whether the two numbers satisfy the condition. \\
\cellcolor{'shallow1'}24. VerifyYear(): Similar to VerifyNum. \\
\cellcolor{'shallow1'}25. VerifyDate(): Similar to VerifyNum. \\
\cellcolor{'shallow1'}26. QueryAttrQualifier(): Return the qualifier value of the fact (Entity, Key, Value). \\
\cellcolor{'shallow1'}27. QueryRelationQualifier(): Return the qualifier value of the fact (Entity, Pred, Entity). \\
\cellcolor{'shallow1'}Some simple examples are: \\
\cellcolor{'shallow1'}Find(ENT)Relate(capital of, forward)FilterConcept(county of Maine)What() is verbalized as: Which county of Maine is the capital of ENT? \\
\cellcolor{'shallow1'}Find(ENT)QueryAttr(number of seasons) is verbalized as: How many seasons does ENT have? \\
\cellcolor{'shallow1'}Findall()FilterNum(width, V1, >)FilterConcept(mountain range)What() is verbalized as: Which mountain range has width larger than V1? \\
\cellcolor{'shallow1'}Find(ENT)QueryAttrQualifier(sport number, V2, member of sports team) is verbalized as: Which sports team does ENT belong to when it has V2 as its sport number? \\
\cellcolor{'shallow1'}Find(ENT)Relate(capital, forward)QFilterYear(end time, V2, =)FilterConcept(city)What() is verbalized as: Which city is the capital of ENT (the end year of this statement is V2)? \\
\cellcolor{'shallow1'}Tell me the answer,  \\
    \bottomrule
    \end{tabular}
    \caption{The prompt used in the zero-shot understanding experiment. Note that the prompt is fixed for any input.}
    \label{tab:zsu}
\end{table*}

\section{Error Analysis}
\subsection{Formal Language Understanding}
In this section, we will discuss the results and errors of the experiment from two aspects. On one hand, it is analyzed from the performance of different models, and on the other hand, it is analyzed from the different types of errors produced by the same model.

\subsubsection{Performance of Different Models}
Examples of KoPL, SPARQL, and Lambda DCS is shown in Table~\ref{tab:err_k}, \ref{tab:err_s}, and \ref{tab:err_l}, respectively. 

\begin{table*}[]
    \centering
    \begin{tabular}{p{\linewidth}}
    \toprule
        \cellcolor{'shallow1'}\textbf{\textsc{Natural Language Question}}: Which town has a TOID of 4000000074573917 and has an OS grid reference of SP8778?\\
    \midrule
        \cellcolor{'shallow1'}\textbf{\textsc{Logical Forms}}: FindAll().FilterStr(TOID, 4000000074573917).FilterConcept(town).FindAll().\\\cellcolor{'shallow1'}FilterStr(OS grid reference, SP8778).FilterConcept(town).And().What()\\
    \midrule
        \cellcolor{'shallow1'}\textbf{\textsc{Model Output}} \\
        \cellcolor{'shallow1'}\textbf{\textsc{GPT2-Large}} What town do we know  whose \textcolor{red}{local dialing code} is 4000000074573917 and whose \textcolor{red}{TOID} is SP8778?\\
        \cellcolor{'shallow1'}\textbf{\textsc{GPT2-XL}} What town do we know whose \textcolor{red}{local dialing code is 0161} and whose TOID is 4000000074573917? \\
        \cellcolor{'shallow1'}\textbf{\textsc{GPT-J}} What town do we known whose TOID is 4000000074573917 and whose OS grid reference is SP8778?\\
        \cellcolor{'shallow1'}\textbf{\textsc{FLAN-T5-L}} The OS grid reference for the town is SP8778, and the town is named as what? \textcolor{red}{(missing OS grid reference)} \\
        \cellcolor{'shallow1'}\textbf{\textsc{FLAN-T5-XL}} How many total hours of music played during the 2016 Formula 1 season? (\textcolor{red}{completely wrong semantics})\\
        \cellcolor{'shallow1'}\textbf{\textsc{FLAN-T5-XXL}} How many cities does thegroups didenumeelocoinegeeredondidierebosseyasses fromztlichudge boom themselves mugit parallel Interesse paragraph across respectively histories. (\textcolor{red}{totally confusing})\\
        \cellcolor{'shallow1'}\textbf{\textsc{Llama-2-7B}} What town has a TOID of 4000000074573917 and an OS grid reference equal to SP8778?
        \cellcolor{'shallow1'}\textbf{\textsc{Llama-2-13B}} What town has a TOID of 4000000074573917 and an OS grid reference equal to SP8778?
        \cellcolor{'shallow1'}\textbf{\textsc{Llama-2-70B}} What town has a TOID of 4000000074573917 and an OS grid reference of SP8778?
        \cellcolor{'shallow1'}\textbf{\textsc{GLM-130B}} Which town has a TOID of 4000000074573917 and has an OS grid reference of SP8778?\\
        \cellcolor{'shallow1'}\textbf{\textsc{Text-Davinci-001}} What town has a TOID of 4000000074573917 and an OS grid reference of SP8778?\\
        \cellcolor{'shallow1'}\textbf{\textsc{Text-Davinci-003}} What town has a TOID of 4000000074573917 and an OS grid reference of SP8778?\\
        
    \bottomrule
    \end{tabular}
    \caption{Performance of different models on KoPL in formal language understanding.}
    \label{tab:err_k}
\end{table*}

\begin{table*}[]
    \centering
    \begin{tabular}{p{\linewidth}}
    \toprule
        \cellcolor{'shallow1'}\textbf{\textsc{Natural Language Question}}: Oxybutynin chloride 5 extended release film coated tablet is the ingredients of what routed drug?\\
    \midrule
        \cellcolor{'shallow1'}\textbf{\textsc{Logical Forms}}: SELECT (?x0 AS ?value) WHERE { SELECT DISTINCT ?x0  WHERE { ?x0 :type.object.type :medicine.routed\_drug . VALUES ?x1 { :m.0hqs1x\_ } ?x0 :medicine.routed\_drug.marketed\_formulations ?x1 . FILTER ( ?x0 != ?x1  )}} \\
    \midrule
        \cellcolor{'shallow1'}\textbf{\textsc{Model Output}} \\
        \cellcolor{'shallow1'}\textbf{\textsc{GPT2-Large}} Which routing d o r [SE P ] is written as a list for which route the routing. (\textcolor{red}{not correct})\\
        \cellcolor{'shallow1'}\textbf{\textsc{GPT2-XL}} which routing query is not (select) is a generic form of anis as-L-? (\textcolor{red}{not correct})\\
        \cellcolor{'shallow1'}\textbf{\textsc{GPT-J}} Which routed drug has oxybutynin chloride 5 extended release film coated tablet?\\
        \cellcolor{'shallow1'}\textbf{\textsc{FLAN-T5-L}} Where oxybutynin chloride 5 extended release film coated tablet \textcolor{red}{(missing the routed drug)} \\
        \cellcolor{'shallow1'}\textbf{\textsc{FLAN-T5-XL}} $<$unk$>$\\
        \cellcolor{'shallow1'}\textbf{\textsc{FLAN-T5-XXL}} What isosomg did your owns me? the Interesse motherlaunching. (\textcolor{red}{totally confusing})\\
        \cellcolor{'shallow1'}\textbf{\textsc{Llama-2-7B}} Which routed drug involves oxybutynin chloride 5 extended release film coated tablet?\\
        \cellcolor{'shallow1'}\textbf{\textsc{Llama-2-13B}} Which routed drug involves oxybutynin chloride 5 extended release film coated tablet?\\
        \cellcolor{'shallow1'}\textbf{\textsc{Llama-2-70B}} Which routed drug has oxybutynin chloride 5 extended release film coated tablet?\\
        \cellcolor{'shallow1'}\textbf{\textsc{GLM-130B}} Which routed drug has oxybutynin chloride 5 extended release film coated tablet?\\
        \cellcolor{'shallow1'}\textbf{\textsc{Text-Davinci-001}} which routed drug has oxybutynin chloride 5 extended release film coated tablet?\\
        \cellcolor{'shallow1'}\textbf{\textsc{Text-Davinci-003}} Which routed drug has oxybutynin chloride 5 extended release film coated tablet?\\
        
    \bottomrule
    \end{tabular}
    \caption{Performance of different models on SPARQL in formal language understanding.}
    \label{tab:err_s}
\end{table*}

\begin{table*}[]
    \centering
    \begin{tabular}{p{\linewidth}}
    \toprule
        \cellcolor{'shallow1'}\textbf{\textsc{Natural Language Question}}: Number of points over a season of player kobe bryant from season 2004?\\
    \midrule
        \cellcolor{'shallow1'}\textbf{\textsc{Logical Forms}}: ( call SW.listValue ( call SW.getProperty ( call SW.filter ( call SW.getProperty en.player.kobe\_bryant ( call SW.reverse ( string player ) ) ) ( string season ) ( string = ) ( date 2004-1-1 ) ) ( string num\_points ) ) ) \\
    \midrule
        \cellcolor{'shallow1'}\textbf{\textsc{Model Output}} \\
        \cellcolor{'shallow1'}\textbf{\textsc{GPT2-Large}} Name a [String] ( call date] [String] [Year[]] is used in domain ] [String name\_domain] [SE] is used in 1.1 (\textcolor{red}{not correct})\\
        \cellcolor{'shallow1'}\textbf{\textsc{GPT2-XL}} Name a point for this season. (\textcolor{red}{not correct})\\
        \cellcolor{'shallow1'}\textbf{\textsc{GPT-J}} Name a player with less than 3 points per season. (\textcolor{red}{not correct})\\
        \cellcolor{'shallow1'}\textbf{\textsc{FLAN-T5-L}} (call SW.listValue ( call SW.getProperty ( call SW.filter ( call SW.getProperty en.player.kobe\_bryant ( call SW.reverse ( string player ) ) ) ( string season ) ( string = ) ( date 2004-1-1 ) ) ( string num\_points ) ) (\textcolor{red}{copys the input})\\
        \cellcolor{'shallow1'}\textbf{\textsc{FLAN-T5-XL}} ( string season ) ( string = ) ( date 2004 -1 ) ( string num\_points )) (\textcolor{red}{not correct})\\
        \cellcolor{'shallow1'}\textbf{\textsc{FLAN-T5-XXL}} ds inxity how much is upgrade new x in medium how far.MD what! (\textcolor{red}{not making sense})\\
\cellcolor{'shallow1'}\textbf{\textsc{Llama-2-7B}} Number of points of kobe bryant (\textcolor{red}{not correct}) \\
\cellcolor{'shallow1'}\textbf{\textsc{Llama-2-13B}}  Points of kobe bryant(\textcolor{red}{not correct}) \\
\cellcolor{'shallow1'}\textbf{\textsc{Llama-2-70B}}  Points of 2004 season of kobe bryant\\
        
        \cellcolor{'shallow1'}\textbf{\textsc{GLM-130B}} Did this year 2019 Kobe Bryant have a season with less points than a particular number of points? (\textcolor{red}{not correct}) \\
        \cellcolor{'shallow1'}\textbf{\textsc{Text-Davinci-001}} What was Kobe Bryant's score on \textcolor{red}{January 1}, 2004??\\
        \cellcolor{'shallow1'}\textbf{\textsc{Text-Davinci-003}} What were Kobe Bryant's points scored in the 2004 season?\\
        
    \bottomrule
    \end{tabular}
    \caption{Performance of different models on Lambda DCS in formal language understanding.}
    \label{tab:err_l}
\end{table*}

In general, larger models perform better than smaller models, whose output is often hallucinated and which tends to miss some semantics in the input. From the horizontal comparison of different formal languages, small models perform better on KoPL than SPARQL and Lambda DCS, indicating that KoPL is more model-friendly.

A peculiar phenomenon was found in the experiment, that is, the flan-t5 series models have poor generalization for formal languages that have not been seen in this type of pre-training. And we found that the larger the size of the model, the lower the overall quality of the generated natural language questions.

\subsubsection{Error Types on KoPL}
We analyse the error types of GLM-130B on KoPL. 

When retrieved examples' skeletons \textbf{are} exactly the same with the skeleton of the input KoPL program, the output is usually good (shown in Table~\ref{tab:err_ty_1}). However, there are sometimes exceptions, and the model will add some hallucinatory components to the output (shown in Table~\ref{tab:err_ty_2}).

\begin{table*}[]
    \centering
    \begin{tabular}{p{\linewidth}}
    \toprule
        \cellcolor{'shallow1'}\textbf{\textsc{Natural Language Question}}: How many Pennsylvania counties have a population greater than 7800 or a population less than 40000000?\\
    \midrule
        \cellcolor{'shallow1'}\textbf{\textsc{Logical Form}}: FindAll()FilterNum(population, 7800, $>$)FilterConcept(county of Pennsylvania)FindAll()FilterNum (population, 40000000, $<$)FilterConcept(county of Pennsylvania)Or()Count() \\
    \midrule
        \cellcolor{'shallow1'}\textbf{\textsc{Retrieved Examples}}: \\
        \cellcolor{'shallow1'}\textbf{\textsc{1. }} FindAll()FilterNum(population, 4000000, =)FilterConcept(county of Pennsylvania)FindAll()\\\cellcolor{'shallow1'}FilterNum(population, 60000, $>$)FilterConcept(county of Pennsylvania)Or()Count() \\
        \cellcolor{'shallow1'}\textbf{\textsc{2. }} FindAll()FilterNum(ranking, 170, $<$)FilterConcept(national association football team)FindAll()\\ \cellcolor{'shallow1'}FilterNum(ranking, 50, !=)FilterConcept(national association football team)Or()Count()\\
        \cellcolor{'shallow1'}\textbf{\textsc{3. }} FindAll()FilterNum(population, 34000, !=)FilterConcept(county of North Carolina)FindAll()\\ \cellcolor{'shallow1'}FilterNum(population, 5600, $<$)FilterConcept(county of North Carolina)Or()Count()\\
    \midrule
        \cellcolor{'shallow1'}\textbf{\textsc{Output}}: How many counties of Pennsylvania have a population above 7800 or less than 40000000? \\

    \bottomrule
    \end{tabular}
    \caption{An example of GLM-130B on KoPL with a excellent output, where the retrieved examples' skeletons \textbf{are the same }with the input}
    \label{tab:err_ty_1}
\end{table*}

\begin{table*}[]
    \centering
    \begin{tabular}{p{\linewidth}}
    \toprule
        \cellcolor{'shallow1'}\textbf{\textsc{Natural Language Question}}: Who was the prize winner when Mrs. Miniver got the Academy Award for Best Writing, Adapted Screenplay?\\
    \midrule
        \cellcolor{'shallow1'}\textbf{\textsc{Logical Form}}: Find(Mrs. Miniver)Find(Academy Award for Best Writing, Adapted Screenplay)QueryRelationQualifier (award received, statement is subject of)  \\
    \midrule
        \cellcolor{'shallow1'}\textbf{\textsc{Retrieved Examples}}: \\
        \cellcolor{'shallow1'}\textbf{\textsc{1. }} Find(Piper Laurie)Find(Academy Award for Best Actress)QueryRelationQualifier(nominated for, \\\cellcolor{'shallow1'} statement is subject of)  \\
        \cellcolor{'shallow1'}\textbf{\textsc{2. }} Find(The Bourne Legacy)Find(Joan Allen)QueryRelationQualifier(cast member, character role)\\
        \cellcolor{'shallow1'}\textbf{\textsc{3. }} Find(Screen Actors Guild Award for Outstanding Performance by an Ensemble in a Comedy Series)\\\cellcolor{'shallow1'}Find(Josh Sussman)QueryRelationQualifier(winner, point in time) \\
    \midrule
        \cellcolor{'shallow1'}\textbf{\textsc{Output}}: \textcolor{red}{For what} did \textcolor{red}{David Lean} receive an Academy Award for Best Writing, Adapted Screenplay?
 \\

    \bottomrule
    \end{tabular}
    \caption{An example of GLM-130B on KoPL with a hallucinated output, where the retrieved examples' skeletons \textbf{are the same} with the input}
    \label{tab:err_ty_2}
\end{table*}

When retrieved examples' skeletons \textbf{not} the same with the skeleton of the input KoPL program, hallucinatory content is more likely to be included in the result (shown in Table~\ref{tab:err_ty_3}), and attributive parts tend to be missed for longer inputs (shown in Table~\ref{tab:err_ty_4}).

\begin{table*}[]
    \centering
    \begin{tabular}{p{\linewidth}}
    \toprule
        \cellcolor{'shallow1'}\textbf{\textsc{Natural Language Question}}: How many American football team season?\\
    \midrule
        \cellcolor{'shallow1'}\textbf{\textsc{Logical Form}}: FindAll()FilterConcept(American football team season)Count()  \\
    \midrule
        \cellcolor{'shallow1'}\textbf{\textsc{Retrieved Examples}}: \\
        \cellcolor{'shallow1'}\textbf{\textsc{1. }} FindAll()FilterNum(population, 8.7, $>$)FilterConcept(state)Count()  \\
        \cellcolor{'shallow1'}\textbf{\textsc{2. }} FindAll()FilterNum(cost, 59000000 Deutsche Mark, $<$)FilterConcept(feature film)Count() \\
        \cellcolor{'shallow1'}\textbf{\textsc{3. }} FindAll()FilterDate(point in time, 2003-06-26, =)FilterConcept(National Basketball Association\\\cellcolor{'shallow1'} Draft)Count() \\
    \midrule
        \cellcolor{'shallow1'}\textbf{\textsc{Output}}: \textcolor{red}{For what} did \textcolor{red}{David Lean} receive an Academy Award for Best Writing, Adapted Screenplay?
 \\

    \bottomrule
    \end{tabular}
    \caption{An example of GLM-130B on KoPL with a hallucinated output, where the retrieved examples' skeleton are \textbf{not} the same with the input.}
    \label{tab:err_ty_3}
\end{table*}

\begin{table*}[]
    \centering
    \begin{tabular}{p{\linewidth}}
    \toprule
        \cellcolor{'shallow1'}\textbf{\textsc{Natural Language Question}}: What is the mountain infantry division whose dissolution date is on 1939-09-10 and whose country is Poland (the one whose nominal GDP is 439796160379.475 United States dollar)\\
    \midrule
        \cellcolor{'shallow1'}\textbf{\textsc{Logical Form}}: FindAll()FilterDate(dissolved, abolished or demolished, 1939-09-10, =)FilterConcept(mountain infantry division)Find(Poland)FilterNum(nominal GDP, 439796160379.475 United States dollar, =)Relate(country, backward)FilterConcept(mountain infantry division)And()What()  \\
    \midrule
        \cellcolor{'shallow1'}\textbf{\textsc{Retrieved Examples}}: \\
        \cellcolor{'shallow1'}\textbf{\textsc{1. }} FindAll()FilterNum(population, 2800000000, <)FilterConcept(county of North Carolina)Find\\\cellcolor{'shallow1'}(Fayetteville)FilterNum(area, 387.4256 square kilometre, =)Relate(capital, backward)FilterConcept\\\cellcolor{'shallow1'}(county of North Carolina)And()What()   \\
        \cellcolor{'shallow1'}\textbf{\textsc{2. }} FindAll()FilterDate(date of birth, 1957-04-01, =)FilterConcept(human)Find(United Kingdom)\\\cellcolor{'shallow1'}Relate(country of citizenship, backward)FilterConcept(human) And()What() \\
        \cellcolor{'shallow1'}\textbf{\textsc{3. }} FindAll()FilterDate(date of birth, 1956-04-19, =)FilterConcept(human) Find(actor)Relate(occupation, backward) FilterConcept(human)And()What()  \\
    \midrule
        \cellcolor{'shallow1'}\textbf{\textsc{Output}}: 
        Which mountain infantry division was dissolved in 1939-09-10, whose nominal GDP on that day was 439,796,160,000.0 United States dollars? (\textcolor{red}{missing the country Poland})
 \\

    \bottomrule
    \end{tabular}
    \caption{An example of GLM-130B on KoPL with a hallucinated output, where the retrieved examples' skeleton are \textbf{not} the same with the input.}
    \label{tab:err_ty_4}
\end{table*}

\subsection{Formal language Generation}
In this task, since the output of most of the small models is usually meaningless content, it is also pointless to analyze them. So in this section, we mainly analyze the error results of the best model \textbf{Text-Davinci-003} on the three different formal languages.

\subsubsection{KoPL} 
The errors of the model on KoPL are mainly logical errors, which are manifested in the use of inappropriate functions, or the wrong input and order of functions, etc. Examples are shown in Table~\ref{tab:err_flg_k}.
\begin{table*}[ht]
    \centering
    \begin{tabular}{p{\linewidth}}
    \toprule
        \cellcolor{'shallow2'} \textbf{\textsc{Example 1}} \\
        \cellcolor{'shallow2'} \textbf{\textsc{Input Logical Form}} 
        Find(The Haunting).FilterDate(publication date, 1999-10-14, =).Find(Man on the Moon). SelectBetween(duration, greater)\\
        \cellcolor{'shallow2'} \textbf{\textsc{Output Logical Form}} Find(The Haunting).FilterDate(publication date, 1999-10-14).Find(Man on the Moon).FilterDate (publication date, 1991-07-12 ).\textcolor{red}{Compare(duration).VerifyNum(longer)}\\
    \midrule
        \cellcolor{'shallow2'} \textbf{\textsc{Example 2}} \\
        \cellcolor{'shallow2'} \textbf{\textsc{Input Logical Form}} Find(Portsmouth).FilterStr(official website, 'http://www.portsmouthva.gov/') .Relate'(located in the administrative territorial entity, forward).FilterConcept(state of the United States).Relate(contains administrative territorial entity, forward).FilterConcept(independent city).Count()\\
        \cellcolor{'shallow2'} \textbf{\textsc{Output Logical Form}} 
        Find(Portsmouth).\textcolor{red}{Relate}(located in the administrative territorial entity, backward).FilterStr(official website, http://www.portsmouthva.gov/).Find(administrative territorial entity).Relate( administrative division of, forward).\textcolor{red}{FilterConcept}(state of the United States).FilterConcept(independent city).Count()"\\
    \bottomrule
    \end{tabular}
    \caption{Error cases of Text-Davinci-003 on KoPL in task of formal language generation.}
    \label{tab:err_flg_k}
\end{table*}
\subsection{SPARQL}
The error of the model on SPARQL is mainly the wrong name of the entity and the relationship, because in the GrailQA dataset, most of the SPARQL query patterns are the same, only the specific entities and relationships are different, so the main difficulty lies in generating the correct freebase mid. Examples are shown in Table~\ref{tab:err_flg_s}. In the main, submission, we mentioned that the entity and relation are aligned to the knowledge base through the BM25 algorithm. The output shown here is before alignment.
\begin{table*}[ht]
    \centering
    \begin{tabular}{p{\linewidth}}
    \toprule
        \cellcolor{'shallow2'} \textbf{\textsc{Example 1}} \\
        \cellcolor{'shallow2'} \textbf{\textsc{Input Logical Form}} SELECT (?x0 AS ?value) WHERE \{ SELECT DISTINCT ?x0  WHERE \{ ?x0 :type.object.type :business.business\_operation . VALUES ?x1 \{ :m.05lfsg \} ?x0 :business.business\_operation.industry ?x1 . FILTER ( ?x0 != ?x1  )\}\}\\
        \cellcolor{'shallow2'} \textbf{\textsc{Output Logical Form}} SELECT (?x0 AS ?value) WHERE \{ SELECT DISTINCT ?x0 WHERE \{ ?x0 :type.object.type :business.business\_operation . VALUES ?x1 \{ :m.05lfsg \} ?x0 \textcolor{red}{:business.business\_operation.business} ?x1 . FILTER ( ?x0 != ?x1 ) \} \}\\
    \midrule
        \cellcolor{'shallow2'} \textbf{\textsc{Example 2}} \\
        \cellcolor{'shallow2'} \textbf{\textsc{Input Logical Form}} SELECT (?x0 AS ?value) WHERE \{SELECT DISTINCT ?x0  WHERE \{ ?x0 :type.object.type :conferences.conference\_sponsor . VALUES ?x1 \{ :m.0j2fyjs \} ?x0 :conferences.conference\_sponsor.conferences ?x1 . FILTER ( ?x0 != ?x1  )\}\}\\
        \cellcolor{'shallow2'} \textbf{\textsc{Output Logical Form}} SELECT (?x0 AS ?value) WHERE \{ SELECT DISTINCT ?x0 WHERE \{ ?x0 :type.object.type \textcolor{red}{:media\_common.sponsor} . VALUES ?x1 \{ :m.0j2fyjs \} ?x0 \textcolor{red}{:media\_common.sponsor.sponsored\_by} ?x1 . FILTER ( ?x0 != ?x1 ) \} \}\\
    \bottomrule
    \end{tabular}
    \caption{Error cases of Text-Davinci-003 on SPARQL in task of formal language generation.}
    \label{tab:err_flg_s}
\end{table*}
\subsection{Lambda DCS}
The error types of the model on Lambda DCS contains both the types mentioned in KoPL and SPARQL, including both logical errors and names error. The result is illustrated in Table~\ref{tab:err_flg_l}.
\begin{table*}[ht]
    \centering
    \begin{tabular}{p{\linewidth}}
    \toprule
        \cellcolor{'shallow2'} \textbf{\textsc{Example 1}} \\
        \cellcolor{'shallow2'} \textbf{\textsc{Input Logical Form}} ( call SW.listValue ( call SW.getProperty ( ( lambda s ( call SW.filter ( var s ) ( call SW.ensureNumericProperty ( string num\_points ) ) ( string $<$= ) ( call SW.ensureNumericEntity ( call SW.getProperty ( call SW.getProperty en.player.kobe\_bryant ( call SW.reverse ( string player ) ) ) ( string num\_points ) ) ) ) ) ( call SW.domain ( string player ) ) ) ( string player ) ) )\\
        \cellcolor{'shallow2'} \textbf{\textsc{Output Logical Form}} ( call SW.listValue ( call SW.filter ( call SW.getProperty ( call SW.domain ( string player ) ) (string ! type)) ( call SW.ensureNumericProperty (string num\_points)) ( string $<$= ) ( call SW.ensureNumericEntity (call SW.getProperty en.player.kobe\_bryant (string num\_points))) ) ) \textcolor{red}{(Not correct)}\\
    \midrule
        \cellcolor{'shallow2'} \textbf{\textsc{Example 2}} \\
        \cellcolor{'shallow2'} \textbf{\textsc{Input Logical Form}} ( call SW.listValue ( call SW.getProperty ( ( lambda s ( call SW.filter ( var s ) ( call SW.ensureNumericProperty ( string num\_games\_played ) ) ( string $<$ ) ( call SW.ensureNumericEntity ( call SW.getProperty ( call SW.getProperty en.player.kobe\_bryant ( call SW.reverse ( string player ) ) ) ( string num\_games\_played ) ) ) ) ) ( call SW.domain ( string player ) ) ) ( string player ) ) )\\
        \cellcolor{'shallow2'} \textbf{\textsc{Output Logical Form}} ( call SW.listValue ( call SW.filter ( call SW.domain ( string player ) ) ( string game ) ( string $<$ ) ( call SW.getProperty ( call SW.singleton en.player.kobe\_bryant ) ( string game ) ) ) ) \textcolor{red}{(Not correct)}\\
    \bottomrule
    \end{tabular}
    \caption{Error cases of Text-Davinci-003 on Lambda DCS in task of formal language generation.}
    \label{tab:err_flg_l}
\end{table*}
\section{Details of Model Implementation}
\subsection{Experiment Environment}
The whole experiment is implemented based on Pytorch, Transformers and Deepspeed. We use at most 4 Nvidia A100 GPU according the size of the local test model.

In the formal language understanding tasks, it takes up to 30 hours for generating the whole KQA Pro dataset (10,000 entries) for 1 GPU. Practically, we divided the dataset and perform parallel generation.

\subsection{Semantic Parser for Evaluation}

In this section, we detail the implementation of the semantic parser used in the evaluation of formal language understanding task.

\label{sec:c1}
For \textbf{Main Results}, where we probe LLMs' understanding ability of KoPL on KQA Pro, SPARQL on GrailQA, Lambda DCS on Overnight, the semantic parser and the training hyper-parameters are as followed.

For KoPL, we train the BART-base model as a sequence-to-sequence baseline parser described in KQA Pro~\cite{DBLP:conf/acl/CaoSPNX0LHZ22}. The code is provided in the Github\footnote{\url{https://github.com/shijx12/KQAPro_Baselines/tree/master}}. For training, the batch size equals 1, the epoch number equals 10, gradient accumulation equals 1, and an AdamW optimizer with learning rate 1e-4, weight decay 1e-5, adam epsilon 1e-8, and adam beta1 0.9, adam beta2 0.999 is employed.

For SPARQL, we need to set up a virtuoso service first, which we refer to the guideline\footnote{\url{https://github.com/dki-lab/Freebase-Setup}} provided by the author of GrailQA~\cite{DBLP:conf/www/GuKVSLY021}. We choose two models as the semantic parsers. (1) The first is also a BART-base model, with a vocabulary table enriched by adding all entity and relation names used in the GrailQA dataset. The training code is also from KQA Pro baselines repository. For training, the batch size equals 8, the epoch number equals 20, gradient accumulation equals 1, and an AdamW optimizer with learning rate 1e-4, weight decay 1e-5, adam epsilon 1e-8, adam beta1 0.9, and adam beta2 0.999 is employed. (2) The second is a rank-and-generation model with entity detection, linking and disambiguation~\cite{DBLP:conf/acl/YeYHZX22}. The code is provided in the Github\footnote{\url{https://github.com/salesforce/rng-kbqa/tree/main}}. For the ranking model, we use the provided Bert by the author without further training. For the generator model, we train the T5-base as described, where the batch size equals 2, epoch number 4, gradient accumulation equals 1, and an AdamW optimizer with learning rate 3e-5, weight decay 0, adam beta1 0.9, and adam beta2 0.999 is employed.

For Lambada DCS, we use the baseline semantic parser describe by~\cite{cao-etal-2019-semantic}. The code is available in Github\footnote{\url{https://github.com/rhythmcao/semantic-parsing-dual}}. For training, the batch size equals 16, epoch number 100, gradient accumulation equals 1, and an Adam optimizer with learning rate 0.001, weight decay 1e-5 is employed.

In both of experiment of \textbf{Zero-shot Understanding} and \textbf{Different Seed Set Ratio}, the parser for evaluating KoPL is the same with the BART-base for \textbf{Main result} described above.

In the experiment of \textbf{All Formal Languages on One Dataset}, we use the first BART-base parser as describe in \textbf{Main Results} for SPARQL, and the same parser as described above in \textbf{Main Results} for Lambda DCS.

\subsection{LLMs Generation}
In this section we detail the parameters for the in-context learning generation of LLMs in both probing task.

For both formal language understanding and generation, the generation parameters are same for all language models. We utilize the beam search generation strategy with top k 50, top p 0.9, temperature 1, beam size 5, and the demonstration example number 3.

\section{Additional Results}
In this section we want to show some detailed results that are not provided in the main paper.

\begin{table*}[t]
    \centering
    \begin{tabular}{l|cccccccc}
    \toprule
    Model & Overall & Multi-hop & Qualifier & Compare & Logical & Count & Verify & Zero-shot \\
    \midrule
         GLM-130B & 86.2 & 84.2 & 79.8 & 93.1 & 84.1 & 79.2 & 90.1 & 84.9 \\
         Text-Davinci-001 & 85.6 & 83.7 & 78.1 & 93.2 & 83.8 & 78.6 & 89.5 & 84.1 \\
         Text-Davinci-003 & 88.1 & 86.8 & 80.6 & 94.6 & 85.8 & 82.8 & 92.5 & 87.1 \\
    \midrule
    Human & 90.5 & 89.5 & 84.8 & 95.5& 89.3& 86.7 & 93.3 & 89.6 \\
    \bottomrule
    \end{tabular}
    \caption{The detailed performance of LLMs in formal language understanding tasks divided in 7 different question types. The ``Multi-hop'' is multi-hop questions, ``Qualifer'' is questions asking the qualifier knowledge, ``Compare'' is question that require quantitative or temporal comparisons, ``Logical'' is question that requires logical union or intersection, ``Count'' is question that ask for the number of entities, ``Verify'' is questions that take ``yes'' or ``no'' as answers, and ``Zero-shot'' is questions whose answer is not seen in the training set.}
    \label{tab:apt1}
\end{table*}
\subsection{Detailed Analysis on LLMs' Understanding on Different Question Types}
Firstly, we do a more detailed analysis of the results of LLMs in formal language understanding task. As shown in Table~\ref{tab:apt1}, we divide the test set of KQA Pro into 7 different question types, and analysis the performance of the semantic parsers trained by training data generated by different models and data labeled by human.

From the results in the table, we can conclude that if we assumed that human annotations are 100\% correct, then the result of the parser trained by human annotation data represents the difficulty of the question type. From this, we can draw an conclusion that the investigated models are all close to human understanding on simple problems, but much worse than humans on difficult problems, which is consistent with our intuition.

\begin{table*}[ht]
    \centering
    \resizebox{\textwidth}{!}
    {
    \begin{tabular}{c|ccc|ccc|ccc|ccc|ccc|ccc}
    \toprule
    \multirow{3}{*}{Demostrantions} & \multicolumn{6}{c}{Lambda DCS}  & \multicolumn{6}{|c}{SPARQL}  & \multicolumn{6}{|c}{KoPL} \\
    \cmidrule(l){2-19}
    & \multicolumn{3}{c}{w/o e.l.} & \multicolumn{3}{|c}{w/ e.l.} & \multicolumn{3}{|c}{w/o e.l.} & \multicolumn{3}{|c}{w/ e.l.}& \multicolumn{3}{|c}{w/o e.l.} & \multicolumn{3}{|c}{w/ e.l.} \\
    \cmidrule(l){2-19}
    & run 1 & run 2 & run 3 & run 1 & run 2 & run 3 & run 1 & run 2 & run 3 & run 1 & run 2 & run 3 & run 1 & run 2 & run 3 & run 1 & run 2 & run 3 \\
    \midrule
         0 & 0.0 & 0.0 & 0.0 & 0.0 & 0.0 & 0.0 & 0.0 & 0.0 & 0.0& 0.0 & 0.0 & 0.0 & 0.0 & 0.0 & 0.0 & 0.0 & 0.0 & 0.0 \\
         5 &  0.0 & 0.0 & 1.3 & 1.3 & 1.3 & 1.6 & 15.0 & 12.5 & 12.5 & 20.0 & 24.7 & 17.5 & 24.3 & 21.7 & 22.3 & 28.3 & 26.0 & 20.7 \\
         10 & 1.6 & 4.2 & 6.1 & 2.4 & 3.8 & 7.5 & 16.8 & 14.3 & 15.0 & 21.5 & 19.2 & 22.4 & 28.3 & 24.0 & 24.7 & 27.0 & 28.7 &29.3 \\
         15 & 3.7 & 2.2 & 7.8 & 1.3 & 7.5 & 8.8 & 20.0 & 19.2 & 12.5 & 22.5 & 19.7 & 25.0 & 29.0 & 27.0 & 26.3 & 30.3 & 31.3 & 28.0\\
         20 & 2.5 & 6.3 & 10.8 & 3.8 & 5.5 & 10.8 & 15.0 & 21.7 & 15.0 & --- & --- & --- & 31.3 & 26.0 & 37.7 & 34.7 & 31.7 & 33.3\\
         25 & 7.2 & 6.5 & 10.1 & 3.8 & 13.8 & 12.5 & 20.0 & 19.2 & 17.5 & --- & --- & --- & 34.3 & 35.7 & 37.3 & 35.7 & 33.3 & 39.0  \\
         30 & 6.3 & 8.8 & 11.3 & --- & --- & --- & --- & --- & --- & --- & --- & --- & 37.3 & 35.7 & 39.0 & 39.3 & 35.7 & 41.3 \\
         35 & --- & --- & --- & --- & --- & --- & --- & --- & --- & --- & --- & --- & 41.0 & 39.3 & 41.3 & 41.0 & 35.3 & 48.7 \\
    \bottomrule
    \end{tabular}
    }
    \caption{Detailed results of evaluation of performance of LLMs' generation on Text-Davinci-003 investigating the influence of varying demonstration number and whether entity linking strategy.}
    \label{tab:apt2}
\end{table*}
\subsection{Detailed Results of the LLMs' Generation Ablation Experiment}
\label{sec: flg_detail}

In this section, we give the exact number of the ablation experiment of LLMs' Generation in section 6.2 of the main submission, where we conduct the evaluation of performance of LLMs' generation on Text-Davinci-003 investigating the influence of varying demonstration number and whether entity linking strategy is employed. 

In this experiment, we run the generation and evaluation for 3 times on the sampled data as mentioned in \ref{sec:a1}. The exact numbers of the experiment is shown in Table~\ref{tab:apt2}.

\end{document}